\documentclass[twoside,11pt]{article}

\usepackage{blindtext}

\usepackage{jmlr2e}

\usepackage{amsmath,amsfonts,bm}

\def\eqref#1{equation~\ref{#1}}

\def\1{\bm{1}}

\DeclareMathAlphabet{\mathsfit}{\encodingdefault}{\sfdefault}{m}{sl}
\SetMathAlphabet{\mathsfit}{bold}{\encodingdefault}{\sfdefault}{bx}{n}

 \usepackage[utf8]{inputenc} \usepackage[T1]{fontenc}    \usepackage{hyperref}       \usepackage{url}            \usepackage{booktabs}       \usepackage{amsfonts}       \usepackage{nicefrac}       \usepackage{microtype}      \usepackage{xcolor}         \usepackage{enumitem}
\usepackage{multirow}
\usepackage{array}
\usepackage{makecell}
\usepackage{tablefootnote}
\usepackage{subcaption}
\usepackage{graphicx}
\usepackage{placeins}
\usepackage{soul}
\usepackage{float}
\usepackage{listings}       \usepackage{algorithm}
\usepackage{algpseudocode}

\usepackage{lastpage}
\jmlrheading{26}{2025}{1-\pageref{LastPage}}{5/25; Revised 11/25}{11/25}{25-1243}{Anke Tang, Li Shen, Yong Luo, Enneng Yang, Han Hu, Lefei Zhang, Bo Du, Dacheng Tao}

\ShortHeadings{FusionBench}{Tang, Shen, Luo, Yang, Hu, Zhang, Du, Tao}
\firstpageno{1}

\usepackage{titletoc}
\usepackage{tocloft}
       \newcommand{\revised}[1]{#1}

\begin{document}

\title{FusionBench: A Unified Library and Comprehensive Benchmark for Deep Model Fusion}

\author{\name Anke Tang$^{1}$ \email anketang@whu.edu.cn
  \AND
  \name Li Shen$^{2\ast}$ \email mathshenli@gmail.com
  \AND
  \name Yong Luo$^{1\ast}$ \email yluo180@gmail.com
  \AND
  \name Enneng Yang$^{2}$ \email ennengyang@gmail.com
  \AND
  \name Han Hu$^{3}$ \email hhu@bit.edu.cn
  \AND
  \name Lefei Zhang$^{1}$ \email zhanglefei@whu.edu.cn
  \AND
  \name Bo Du$^{1}$\thanks{Corresponding authors: Li Shen (mathshenli@gmail.com), Yong Luo (yluo180@gmail.com) and Bo Du (dubo@whu.edu.cn).} \email dubo@whu.edu.cn
  \AND
  \name Dacheng Tao$^{4}$ \email dacheng.tao@ntu.edu.sg\\
  \addr{
    $^1$School of Computer Science, National Engineering Research Center for Multimedia Software and Hubei Key Laboratory of Multimedia and Network Communication Engineering, Wuhan University, Wuhan, China\,
    $^2$Shenzhen Campus of Sun Yat-sen University, China\,
    $^3$Beijing Institute of Technology, Beijing, China\,
    $^4$Nanyang Technological University, Singapore
  }
}

\editor{Zeyi Wen}

\maketitle

\begin{abstract}Deep model fusion is an emerging technique that unifies the predictions or parameters of several deep neural networks into a single better-performing model in a cost-effective and data-efficient manner.
  Although a variety of deep model fusion techniques have been introduced, their evaluations tend to be inconsistent and often inadequate to validate their effectiveness and robustness.
  We present \textit{FusionBench}, the first benchmark and a unified library designed specifically for deep model fusion.
  Our benchmark consists of multiple tasks, each with different settings of models and datasets.
  This variety allows us to compare fusion methods across different scenarios and model scales.
  Additionally, FusionBench serves as a unified library for easy implementation and testing of new fusion techniques.
  FusionBench is open source and actively maintained, with community contributions encouraged.

  \raisebox{-1.5pt}{\includegraphics[height=10pt]{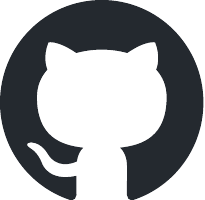}} Code Repository: \url{https://github.com/tanganke/fusion_bench}

  \raisebox{-1.5pt}{\includegraphics[height=10pt]{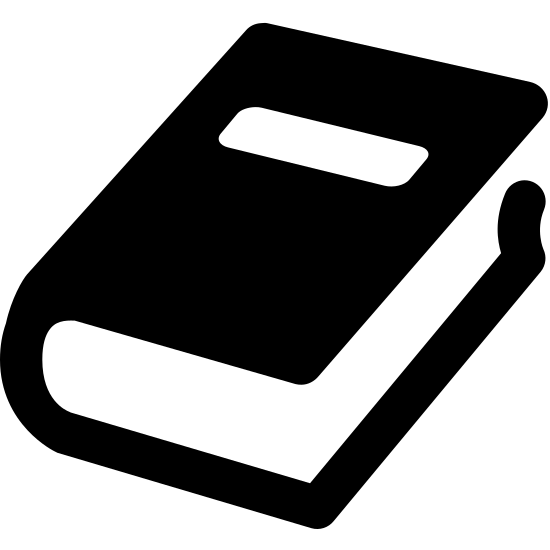}} Online Documentation: \url{https://tanganke.github.io/fusion_bench}

  \raisebox{-1.5pt}{\includegraphics[height=10pt]{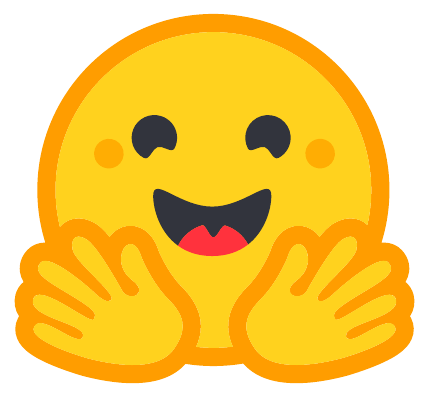}} HuggingFace Collections (models \& datasets): \url{https://huggingface.co/tanganke}
\end{abstract}


\section{Introduction}
\label{section:introduction}

In recent years, a new paradigm called ``learn from model'' has emerged in the field of deep learning, which focuses on leveraging the knowledge embedded in existing models to develop new ones~\citep{zheng2023learn}.
This paradigm has been widely adopted in various scenarios, such as model tuning~\citep{he2022masked,chung2024scaling}, model distillation~\citep{hinton2015distilling}, model pruning~\citep{han2015deep,asif2020ensemble}, model editing~\citep{mitchell2021fast,zhang2024comprehensive}, and so on.
Among these methods, deep model fusion is particularly appealing. It merges the parameters or predictions of multiple models to create a more robust and efficient unified model.
Due to its effectiveness and scalability, many new techniques for deep model fusion have recently been proposed~\citep{li2023deep}.

Deep model fusion offers both scalability and data efficiency by utilizing the knowledge embedded in pre-existing models, rather than requiring training from scratch. This approach significantly accelerates model development, making it a practical solution in the current era dominated by large foundation models.
Despite its potential, the evaluation of deep model fusion techniques often suffers from inconsistency and inadequacy.
Current libraries like MergeKit~\citep{goddard2024arcee} are primarily focused only on large language model merging and do not provide evaluation frameworks.
Standardized assessments across different tasks and models are lacking, making it challenging to evaluate the effectiveness and robustness of deep model fusion techniques.
The potential reasons for this inconsistency include the rapid development of new techniques, the absence of standardized tasks and models, and the variety of settings (such as different fine-tuning strategies).
Additionally, challenges in implementing or replicating prior work contribute to these inconsistencies.
A more detailed comparison with existing model fusion methods and toolkits is provided in Section~\ref{section:related_work}.

\begin{figure}[t]
  \setlength{\abovecaptionskip}{0.1in}
  \centering
  \includegraphics[width=0.86\linewidth]{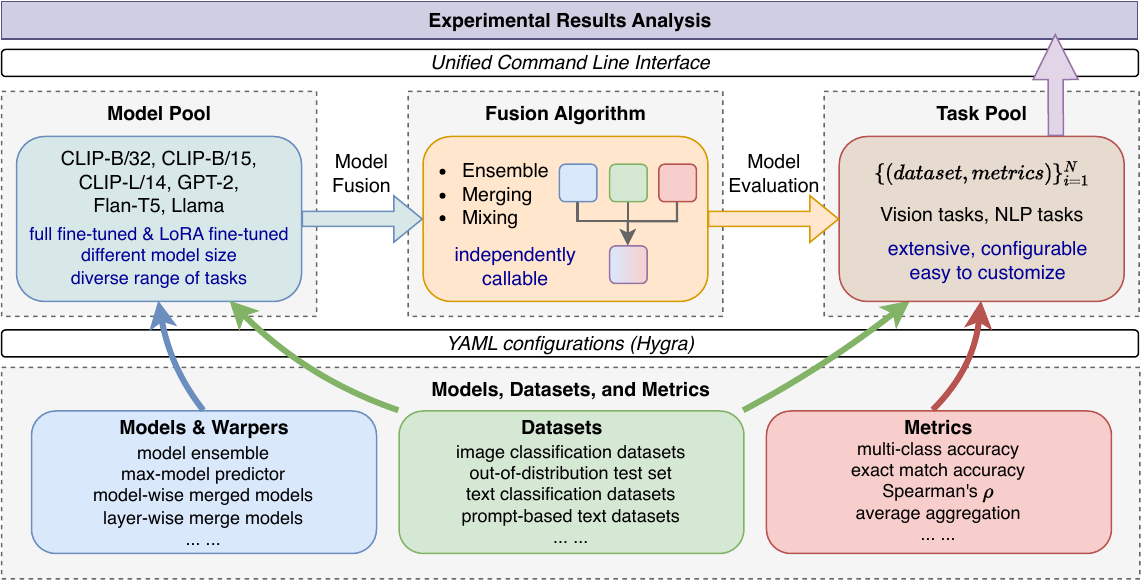}
  \caption{The general framework of the modularized FusionBench codebase.}
  \label{fig:framework_of_model_fusion}
  \vskip -0.16in
\end{figure}

To tackle these challenges, we develop the first comprehensive benchmark dedicated to deep model fusion, called \textit{FusionBench}.
FusionBench is designed as a modular and flexible framework with three primary components: the \textit{Algorithm Module}, the \textit{Model Pool Module}, and the \textit{Task Pool Module} as shown in Figure~\ref{fig:framework_of_model_fusion}.
Each component offers full configurability, enabling straightforward integration of novel algorithms, models, and evaluation tasks.
Through Hydra-based configuration management, all experimental setups are streamlined via a single unified interface.
Furthermore, FusionBench provides extensive documentation, hands-on tutorials, illustrative code samples, and ready-to-use model collections.

\section{FusionBench}
\label{section:benchmark}

FusionBench is available for installation through two simple methods: using pip with `\texttt{pip install fusion-bench}' or cloning the repository directly with `\texttt{git clone}'.
The general framework of the modularized FusionBench codebase is shown in Figure~\ref{fig:framework_of_model_fusion}, which consists of three primary elements: \textit{Algorithm Module}, \textit{Model Pool Module}, and \textit{Task Pool Module}.
In Section~\ref{section:components}, we introduce the three major components, which is designed to be flexible and modular, allowing users to easily run experiments and evaluate the performance of model fusion algorithms.
In Section~\ref{section:implemented_algorithms}, we introduce the implemented model fusion algorithms.
Finally, in Section~\ref{section:documentation_and_tutorials}, we discuss the documentation and tutorials provided to help users understand the benchmark and effectively use the codebase.
In Appendix~\ref{appendix:flowchart}, we provide a flowchart to illustrate the process of running experiments and evaluating the merged models.

\paragraph{Main Program Pipeline.}
The typical pipeline consists of the following steps:
(1) Users configure the experiment by selecting a fusion algorithm, models to merge, and optional evaluation tasks.
(2) The system loads the specified models, algorithm, and evaluation tasks.
(3) The algorithm merges the selected models, with an option to save the resulting model.
(4) When evaluation tasks are specified, the merged model is tested and results are reported.

\subsection{Components}
\label{section:components}

As shown in Figure~\ref{fig:framework_of_model_fusion}, the codebase is composed of three primary components: \textit{Algorithm Module}, \textit{Model Pool Module}, and \textit{Task Pool Module}, which are responsible for implementing the model fusion algorithms, managing the models and datasets used during the algorithm execution, and managing the tasks and evaluation metrics to be evaluated, respectively.
Additionally, we provide a unified command line interface (CLI) `\texttt{fusion\_bench}' to facilitate the use of the codebase and to enable users to easily run experiments.
\begin{itemize}[leftmargin=*,itemsep=0pt]
  \item \textbf{Algorithm Module} is the core component, which contains the implementation of all model fusion algorithms.
        Each algorithm is implemented as a separate Python class inheriting from the base class \texttt{fusion\_bench.BaseAlgorithm}.
        The algorithm classes are flexible - they can be used through our CLI or in custom code.
  \item \textbf{Model Pool Module} provides a unified interface for managing models and datasets.
  \item \textbf{Task Pool Module} evaluates model performance on different tasks, where each task combines a dataset with specific evaluation metrics.
\end{itemize}

\subsection{Implemented Algorithms}
\label{section:implemented_algorithms}

\begin{table}[t]
  \caption{Implemented algorithms in FusionBench.
  }
  \label{table:implemented_algorithms}
  \centering
  \setlength{\tabcolsep}{6pt}
  \fontsize{8}{9}\selectfont
  \begin{tabular}{lc}
    \toprule
    \textbf{Method}                                       & \textbf{Requirement}                        \\
    \midrule
    \multicolumn{2}{c}{\textit{\textbf{Model Ensemble Methods}}}                                        \\
    Simple Ensemble~\citep{sagi2018ensemble}              & -                                           \\
    Weighted Ensemble~\citep{sagi2018ensemble}            & hyperparameter search                       \\
    Max-Model Predictor~\citep{wu2019heterogeneous}       & -                                           \\
    \midrule
    \multicolumn{2}{c}{\textit{\textbf{Model Merging Methods}}}                                         \\
    Simple Average / Modelsoups~\citep{wortsman2022model} & -                                           \\
    Weighted Average~\citep{matena2022merging}            & hyperparameter search                       \\
    Fisher Merging~\citep{matena2022merging}              & compute weights on labeled data             \\
    RegMean~\citep{jin2022dataless}                       & compute weights on labeled data             \\
    RegMean++~\citep{nguyen2025regmean++}                 & compute weights on labeled data             \\
    Concrete Subspace~\citep{tang2023concrete}            & test-time adaptation training               \\
    Task Arithmetic~\citep{ilharco2022editing}            & hyperparameter search                       \\
    Ties-Merging~\citep{yadav2023resolving}               & hyperparameter search                       \\
    Task-Wise AdaMerging~\citep{yang2023adamerging}       & test-time adaptation training               \\
    Layer-Wise AdaMerging~\citep{yang2023adamerging}      & test-time adaptation training               \\
    Representation Surgery~\citep{yang2024representation} & unlabeled data samples                      \\
    TALL mask~\citep{wang2024localizing}                  & -                                           \\
    Task Singular Vectors~\citep{gargiulo2024task}        & -                                           \\
    Isotropic Merging~\citep{marczak2025no}               & -                                           \\
    OPCM~\citep{tang2025merging}                          & -                                           \\
    FW-Merging~\citep{chen2025fw}                         & labeled data samples                        \\
    RanDeS~\citep{zhou2025randes}                         & -                                           \\
    \midrule
    \multicolumn{2}{c}{\textit{\textbf{Model Mixing Methods}}}                                          \\
    MoE-based Upscaling~\citep{komatsuzaki2022sparse}     & pre-training to recover performance         \\
    MoE-based Merging~\citep{komatsuzaki2022sparse}       & training on the combined model              \\
    Depth Upscaling~\citep{kim2023solar}                  & pre-training to recover performance         \\
    Model Recombination~\citep{hu2023fedmr}               & training on the combined model              \\
    Weight-Ensemble MoE~\citep{tang2024merging}           & test-time adaptation training, vision tasks \\
    WE-MoE V2~\citep{shen2024efficient}                   & test-time adaptation training, vision tasks \\
    Pareto-Driven Merging~\citep{tang2024towards}         & training datasets                           \\
    SMILE Upscaling~\citep{tang2024smile}                 & -                                           \\
    \midrule
    \multicolumn{2}{c}{\textit{Model Compression Methods}}                                              \\
    {BitDelta~\citep{liu2024bitdelta}}                    & {calibration data}                          \\
    {Magnitude Pruning~\citep{frankle2020linear}}         & {-}                                         \\
    {Wanda Pruning~\citep{sun2023simple}}                 & {calibration data}                          \\
    {SparseGPT Pruning~\citep{frantar2023sparsegpt}}      & {calibration data}                          \\
    {Expert Pruning~\citep{lu2024not}}                    & {calibration data}                          \\
    \bottomrule
  \end{tabular}
\end{table}

Our benchmark incorporates a comprehensive collection of cutting-edge model fusion algorithms, spanning three key categories: ensemble methods, merging methods, and mixing methods.
Additionally, we have recently extended FusionBench to include model compression methods, which, while not strictly fusion techniques, are complementary approaches that can be applied to reduce the computational overhead of models or compress individual models before and after fusion.
Additionally, we evaluated their prominence in the research literature and their practical utility.
The complete listing of these implemented algorithms can be found in Table~\ref{table:implemented_algorithms} and FusionBench is under active development and will be updated to include the latest and most effective model fusion algorithms.

\subsection{Online Documentation and Tutorials}
\label{section:documentation_and_tutorials}

We provide detailed documentation and tutorials on our project website to help users get started. These resources cover the basics of model fusion, how to run experiments, and how to evaluate results. We also include example pre-run results showing how different fusion algorithms perform across various tasks.

\section{Conclusion and Discussion}
\label{section:conclusion}

FusionBench offers a robust framework for evaluating deep model fusion algorithms and a unified library for developing new ones.
Its scalable and extensible architecture simplifies the creation of deep model fusion techniques.
Furthermore, FusionBench provides curated datasets and models to ensure fair comparisons.
With comprehensive documentation and tutorials, it is accessible to both beginners and experienced researchers.
We encourage the community to use this benchmark to advance the field of deep model fusion.

\section*{Acknowledgments and Disclosure of Funding}
This work is supported by the National Natural Science Foundation of China (Grant No. 62225113, U23A20318, U2336211, 62576364 and 62276195), the Fundamental and Interdisciplinary Disciplines Breakthrough Plan of the Ministry of Education of China (Grant No. JYB2025XDXM704), the Shenzhen Basic Research Project (Natural Science Foundation) Basic Research Key Project (NO. JCYJ20241202124430041), the Foundation for Innovative Research Groups of Hubei Province (Grant No. 2024AFA017) and the Science and Technology Major Project of Hubei Province (Grant No. 2024BAB046). Dr. Tao's research is partially supported by NTU RSR and Start Up Grants. The numerical calculations in this paper have been done on the supercomputing system in the Supercomputing Center of Wuhan University.

\FloatBarrier
\newpage

\appendix

The appendix is organized into several sections, each providing additional insights and details related to different aspects of the main work.\\

\startcontents[sections]
\printcontents[sections]{l}{1}{\setcounter{tocdepth}{2}}  
\vskip 0.2in
\hrule

\FloatBarrier

\section{Related Work of Deep Model Fusion}
\label{section:related_work}

\begin{figure}[tb]
  \centering
  \setlength{\abovecaptionskip}{0.1in}
  \includegraphics[width=0.85\linewidth]{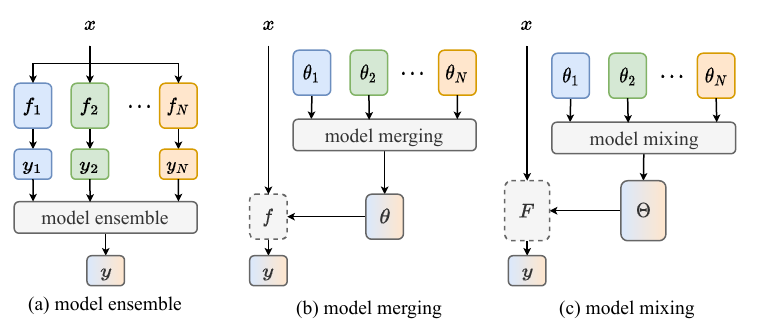}
  \caption{A taxonomy of deep model fusion techniques.}
  \label{fig:model_fusion}
\end{figure}

Deep model fusion, often referred to in the literature simply as ``model fusion'', aims to combine the outputs or parameters of multiple deep neural networks into a unified model with enhanced performance, efficiency, or robustness.
Different researchers may categorize these techniques in various ways based on their understanding and points of view.
Here, we propose a taxonomy that divides these techniques into three major categories: \textit{Model Ensemble}, \textit{Model Merging}, and \textit{Model Mixing}.
Each of these categories approaches model fusion from a unique perspective, offering distinct advantages and applicability.
In the following, we provide detailed explanations, formal definitions, and analyze their strengths and weaknesses.
A visualization of the taxonomy is shown in Figure~\ref{fig:model_fusion}.

\subsection{A Taxonomy of Deep Model Fusion Techniques}

\textbf{Model Ensemble} methods combine the predictions of multiple models to improve the overall performance of a machine learning system~\citep{sagi2018ensemble}, where the collective knowledge is often more accurate and reliable than that of any individual model.
Mathematically, given a set of $N$ models $\{f_1, f_2, \dots, f_N\}$, which can be homogeneous or heterogeneous, we use their predictions to obtain a global prediction $y = \mathcal{A}_{ensemble}(x; f_1, f_2, \dots, f_N; w)$, where $\mathcal{A}_{ensemble}$ is an ensemble algorithm and $w$ are the algorithmic parameters.
Each model $f_i$ can also be associated with a specification to indicate its weight or importance in the ensemble~\citep{pathak2010multiparty,zhouLearnwareFutureMachine2016,wu2021model,tangImprovingHeterogeneousModel2023a}.
Ensemble methods are widely used and effective in improving performance but are often expensive to use and manage.
Recent research has also investigated efficient techniques for model ensembles~\citep{wen2020batchensemble,chen2023split,allingham2021sparse}.

\textbf{Model Merging} methods integrate the parameters of multiple models into a unified model, enhancing efficiency in terms of inference cost and storage, and enabling scalable model fusion.
Given a set of $N$ isomorphic models $\{f_i(\cdot; \theta_i)\}_{i=1}^N$, each parameterized with $\theta_i$, we merge them into a single model with parameters $\theta = \mathcal{A}_{merging}(\theta_1, \theta_2, \dots, \theta_N; w)$, where $\mathcal{A}_{merging}: \mathbb{R}^{N \times d}\rightarrow \mathbb{R}^{d}$ is a merging algorithm and $w$ are the algorithmic parameters. The merged model can be expressed as $f(\cdot; \theta)$.
This method can be implemented through linear interpolation in parameter space~\citep{wortsman2022model,ilharco2022editing,yadav2023resolving,matena2022merging,yu2024language,chronopoulou2023adaptersoup,rame2024rewarded,ortiz2024task,liu2023tangent},
leveraging mode connectivity~\citep{draxler2018essentially,frankle2020linear,benton2021loss,garipov2018loss,qu2024rethink},
aligning features, parameters or gradients~\citep{liu2022deep,ainsworthGitReBasinMerging2023,jin2022dataless,tam2024merging,stoica2023zipit,jang2023personalized,daheim2023model,yang2024representation},
subspace-based methods~\citep{tang2023concrete,wang2024localizing,yi2024safety,zhu2024model,xu2024weight},
and ensemble distillation~\citep{wan2024knowledge,wan2024fusechat}.
Model merging methods are often performed in a data-efficient manner, the algorithmic parameters $w$ can also be learned during test time via test-time adaptation (TTA) training or meta-learning for a more seamless merging~\citep{yang2023adamerging,tang2023concrete}.

\textbf{Model Mixing} methods fuse the components of multiple models to create a new heterogeneous model, which can be more flexible and adaptive than the original models.
Mathematically, given a set of $N$ models $\{f_i(\cdot; \theta_i)\}_{i=1}^N$, each parameterized with $\theta_i \in \mathbb{R}^{d}$, we mix their components to obtain a new model with parameters $\Theta = \mathcal{A}_{mixing}(\theta_1, \theta_2, \dots, \theta_N; w) \in \mathbb{R}^{d'}$, where $\mathcal{A}_{mixing}: \mathbb{R}^{N\times d}\mapsto \mathbb{R}^{d'}$ is a mixing algorithm and $w$ is the algorithmic parameters.
The mixed model can be expressed as $F(\cdot; \Theta)$, which often has more parameters than the original models, and thus can be more expressive and powerful to capture the underlying patterns in the data.
Model mixing methods can be implemented through layer recombinations~\citep{hu2023fedmr,jiang2024evomerge}, model stitching~\citep{lenc2015understanding,moschella2022relative}, or upscale to create a Mixture of Experts (MoE)-based sparse model~\citep{komatsuzaki2022sparse,ye2023taskexpert,tang2024merging,lu2024twin,dai2024deepseekmoe,zhao2024loraretriever,ostapenko2024towards,tang2024smile,yadav2024survey}.

\subsection{Related Benchmarks and Toolkits}

Although several model fusion methods have been proposed, comprehensive benchmarks and unified toolkits are still lacking in this field.
A recent notable work, MergeKit~\citep{goddard2024arcee}, provides a collection of model fusion techniques \textit{specifically designed for merging large language models (LLMs)}, focusing primarily on methods for Transformer-based architectures.
While MergeKit excels in providing practical tools for LLM merging, its scope is inherently limited to a single domain (natural language processing) and a specific model family (Transformers).
Similarly, MergeBench~\citep{he2025mergebench} is targeted exclusively at evaluating LLM merging techniques, providing benchmarks only within the language modeling domain.

In contrast, FusionBench addresses several critical gaps in the current landscape, making FusionBench a more generalized and versatile platform for systematically assessing the performance of model fusion approaches across various domains and architectures:
\textit{(1) Broader Domain Coverage:} Unlike existing tools that focus solely on NLP, FusionBench spans both computer vision and natural language processing domains, enabling cross-domain evaluation and comparison.
\textit{(2) Comprehensive Taxonomy:} While MergeKit and MergeBench concentrate primarily on model merging methods, FusionBench encompasses a complete taxonomy including model ensemble, model merging, and model mixing methods, providing a unified framework for all major fusion paradigms.
\textit{(3) Research-Oriented Evaluation:} FusionBench is designed as a research platform that includes diverse fine-tuned models, standardized evaluation protocols, and comparative analysis utilities across different architectures (CNNs, Vision Transformers, CLIP models, etc.) and tasks.
\textit{(4) Algorithmic Diversity:} Beyond basic merging techniques, FusionBench incorporates advanced methods such as subspace-based fusion, gradient alignment, and mixture-of-experts approaches. \section{Project Structure}
\label{appendix:project_structure}

To provide a clear overview of FusionBench's organization, we present its comprehensive project structure in Listing~\ref{code:project_structure}. This structure reflects the modular design philosophy of our benchmark, with distinct directories for configuration management, documentation, example implementations, core functionality, and testing.

\begin{lstlisting}[language=Bash, caption=Project structure of FusionBench., label=code:project_structure]
|-- config/                   # Configuration files managed by Hydra
|   |-- method/               # Configs for implemented algorithms
|   |-- modelpool/            # Configs for model pool
|   |-- taskpool/             # Configs for task pool
|   |-- model/                # Configs for models
|   |-- dataset/              # Configs for datasets
|-- docs/                     # Documentation (mkdocs)
|-- examples/                 # Example scripts and notebooks
|   |-- {method_name}/        # Method-specific examples
|-- fusion_bench/             # Main package
|   |-- method/               # Fusion algorithm implementations
|   |   |-- {method_name}/    # Method-specific code
|   |-- modelpool/            # Model pool implementations
|   |-- taskpool/             # Task pool implementations
|   |-- ...                   # Models, datasets, and other utilities
|-- tests/                    # Unit tests
\end{lstlisting}
\FloatBarrier
\begin{figure}[hp]
  \centering
  \includegraphics[width=0.75\textwidth]{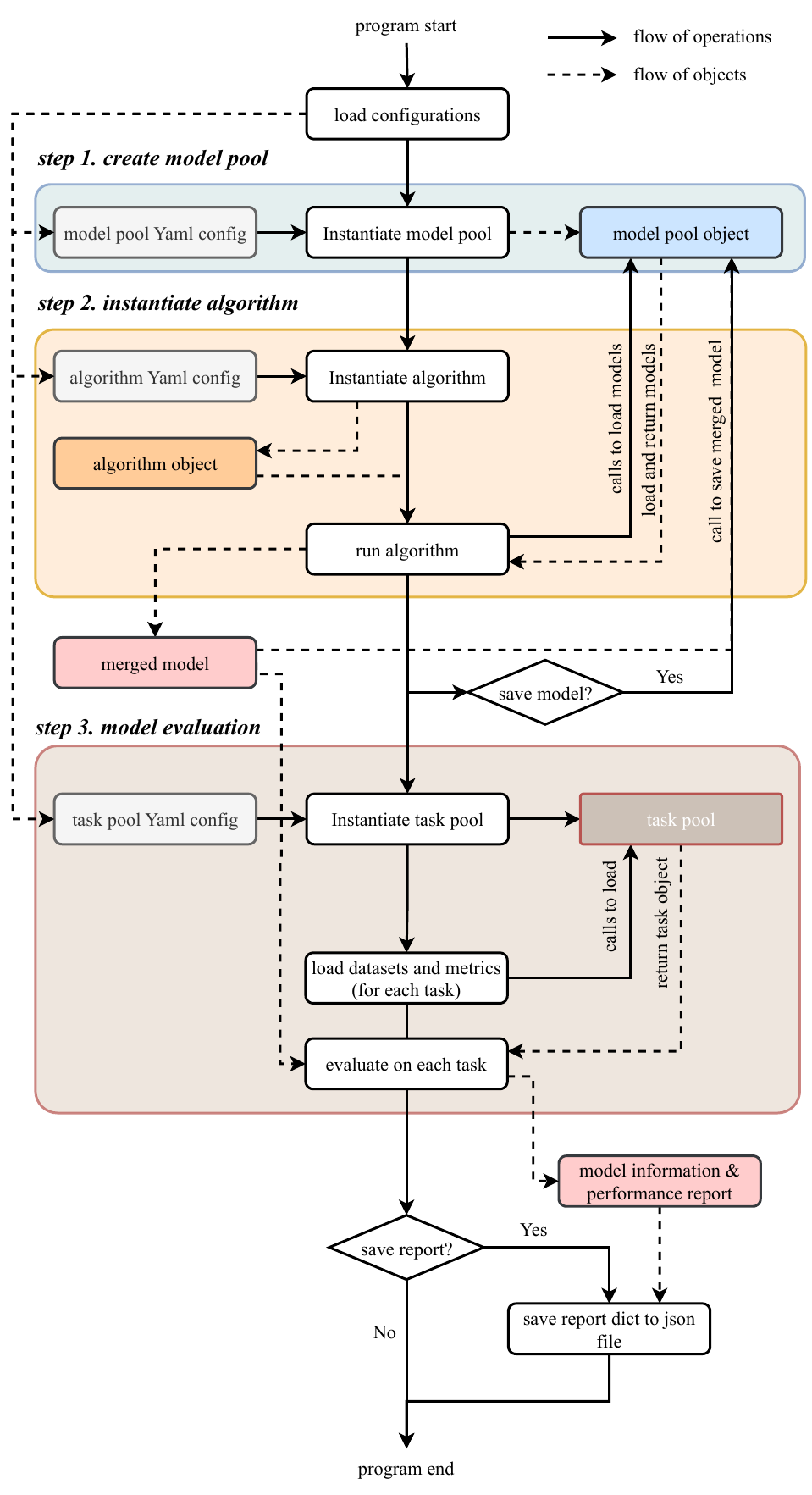}
  \caption{Flowchart of FusionBench.}
  \label{fig:fusion_bench_flow}
\end{figure}

\section{FusionBench Architecture and Implementation Guide}
\label{appendix:flowchart}

Figure~\ref{fig:fusion_bench_flow} illustrates the steps to fuse a model in FusionBench across three main stages: creating a model pool, instantiating and running an algorithm, and (optional) model evaluation.

\paragraph{Pseudocode Example.}
The pseudocode for the pipeline is shown in Algorithm~\ref{alg:main_pipeline}.

\begin{algorithm}[ht]
  \caption{Pseudocode for FusionBench main pipeline}
  \label{alg:main_pipeline}
  \begin{algorithmic}[1]
    \State Load and parse configuration from YAML files
    \State Instantiate \textit{fusion\_algorithm}, \textit{model\_pool}, and (optionally) \textit{task\_pool}
    \State $\textit{merged\_model} \gets \textit{fusion\_algorithm.run}(\textit{model\_pool})$
    \State Save $\textit{merged\_model}$ if save path is specified
    \If{task pool is specified}
    \State $\textit{report} \gets \textit{task\_pool.evaluate}(\textit{merged\_model})$
    \State Save $\textit{report}$ if save path is specified
    \EndIf
  \end{algorithmic}
\end{algorithm}

This design allows users to flexibly experiment with different algorithms, models, and tasks by simply modifying the configuration, without changing the core program logic.

\subsection{Implementing Customized Algorithm in FusionBench}

Here, we provide an example of implementing a customized algorithm in Listing~\ref{code:implementing_new_model_fusion_algorithm}.
The derived algorithm class should inherit from the ``\texttt{BaseAlgorithm}'' class and implement the ``\texttt{run}'' method.
The ``\texttt{run}'' method takes a ``\texttt{BaseModelPool}'' object as input, which is responsible for managing the models and dataset to be loaded.
The ``\texttt{run}'' method should be implemented to fuse the models and return the fused model.

\begin{lstlisting}[language=Python, caption=Implementing new algorithm in FusionBench., label=code:implementing_new_model_fusion_algorithm]
from fusion_bench import BaseAlgorithm, BaseModelPool

class DerivedAlgorithm(BaseAlgorithm):
    """
    An example of a derived algorithm class.
    """

    # _config_mapping maps the attribution to the corresponding key in the configuration file.
    # this is optional and can be used to serialize the object to a configuration file.
    # `self.config.hyperparam_1` will be mapped to the attribute `hyperparam_attr_1`.
    _config_mapping = BaseModelFusionAlgorithm._config_mapping | {
        "hyperparam_attr_1": "hyperparam_1",
        "hyperparam_attr_2": "hyperparam_2",
    }

    def __init__(self, hyperparam_1, hyperparam_2, **kwargs):
        super().__init__(**kwargs)
        self.hyperparam_attr_1 = hyperparam_1
        self.hyperparam_attr_2 = hyperparam_2

    def run(self, modelpool: BaseModelPool):
        # modelpool is an object that responsible for managing the models and dataset to be loaded.
        # implement the fusion algorithm here.
        raise NotImplementedError(
            "DerivedAlgorithm.run() is not implemented."
        )
\end{lstlisting}

As an illustration, we can implement a deep model fusion algorithm that merges models through parameter averaging~\citep{wortsman2022model}. The following code snippet demonstrates this approach (Listing~\ref{code:simple_average_algorithm}):

\begin{lstlisting}[language=Python, caption=Implementation of simple average algorithm in FusionBench., label=code:simple_average_algorithm]
class SimpleAverageAlgorithm(
    BaseAlgorithm,
    SimpleProfilerMixin,
):
    @torch.no_grad()
    def run(self, modelpool: Union[BaseModelPool, Dict[str, nn.Module]]):
        """
        Fuse the models in the given model pool using simple averaging.

        This method iterates over the names of the models in the model pool, loads each model, and appends it to a list.
        It then returns the simple average of the models in the list.

        Args:
            modelpool: The pool of models to fuse.

        Returns:
            The fused model obtained by simple averaging.
        """
        if isinstance(modelpool, dict):
            modelpool = BaseModelPool(modelpool)

        log.info(
            f"Fusing models using simple average on {len(modelpool.model_names)} models."
            f"models: {modelpool.model_names}"
        )
        sd: Optional[StateDictType] = None
        forward_model = None
        merged_model_names = []

        for model_name in modelpool.model_names:
            with self.profile("load model"):
                model = modelpool.load_model(model_name)
                merged_model_names.append(model_name)
                print(f"load model of type: {type(model).__name__}")
            with self.profile("merge weights"):
                if sd is None:
                    # Initialize the state dictionary with the first model's state dictionary
                    sd = model.state_dict(keep_vars=True)
                    forward_model = model
                else:
                    # Add the current model's state dictionary to the accumulated state dictionary
                    sd = state_dict_add(sd, model.state_dict(keep_vars=True))
        with self.profile("merge weights"):
            # Divide the accumulated state dictionary by the number of models to get the average
            sd = state_dict_div(sd, len(modelpool.model_names))

        forward_model.load_state_dict(sd)
        # print profile report and log the merged models
        self.print_profile_summary()
        log.info(f"merged {len(merged_model_names)} models:")
        for model_name in merged_model_names:
            log.info(f"  - {model_name}")
        return forward_model
\end{lstlisting}

\subsection{\revised{Customizing Model Pools in FusionBench}}

\revised{
  The \textit{Model Pool Module} serves as an abstraction layer that manages model loading, saving, and dataset handling during fusion operations.
  To extend FusionBench to new model architectures, users need to implement a custom model pool class that inherits from \texttt{BaseModelPool} and defines architecture-specific operations.
  This design principle ensures that fusion algorithms remain architecture-agnostic and can be applied universally across different model types without modification.

  \textbf{Extensibility to Modern Architectures.}
  FusionBench's modular architecture facilitates straightforward integration of emerging model architectures.
  The key advantage of this design is that once a model pool class is implemented for a particular architecture, all existing fusion algorithms in the benchmark become immediately applicable to that architecture.
  This extensibility has enabled us to expand beyond our initial focus on established models (CLIP-ViT, ResNet-50, GPT-2, and Flan-T5) to encompass state-of-the-art architectures across multiple domains.

  For computer vision tasks, we have integrated modern architectures including ConvNeXt~\citep{liu2022convnet} and DINOv2~\citep{oquab2024dinov}, both of which represent significant advances in efficient visual representation learning.
  These models can be loaded through dedicated model pool classes (\texttt{ConvNeXtForImageClassificationPool} and \texttt{DINOv2ForImageCla-\\ssificationPool}) with comprehensive online documentation available on our project website.

  For natural language processing tasks, our \texttt{CausalLMPool} class provides broad compatibility with contemporary large language model architectures.
  The \texttt{CausalLMPool} leverages the HuggingFace Transformers library, allowing seamless integration of new architectures as they are released.
}
\FloatBarrier
\section{Fine-Tuned Model Collections in FusionBench}
\label{appendix:single_task_results}

This section outlines the comprehensive model collections provided in FusionBench, including their experimental configurations and performance evaluation results across various single-task specializations.

\begin{enumerate}[itemsep=0pt, topsep=0pt, parsep=0pt]
  \item \textbf{CLIP-ViT Models}:
        In FusionBench, we provide comprehensive CLIP-ViT model collections fine-tuned on up to 20 image classification tasks.
        The most commonly used eight tasks are SUN397~\citep{xiao2010sun}, Cars~\citep{krause20133d}, RESISC45~\citep{cheng2017remote}, EuroSAT~\citep{helber2018introducing}, SVHN~\citep{netzer2011reading}, GTSRB~\citep{stallkamp2012man}, MNIST~\citep{lecunGradientbasedLearningApplied1998}, and DTD~\citep{cimpoi2014describing}.
        Additionally, we extend our collection to include 12 more tasks: Oxford Flowers102~\citep{nilsback2008automated}, PatchCamelyon~\citep{veeling2018rotation}, FER2013~\citep{goodfellow2013challenges}, Oxford-IIIT Pet~\citep{parkhi2012cats}, STL-10~\citep{coates2011analysis}, CIFAR-100~\citep{krizhevsky2009learning}, CIFAR-10, Food-101~\citep{bossard2014food}, Fashion-MNIST~\citep{xiao2017fashion}, EMNIST Letters~\citep{cohen2017emnist}, Kuzushiji-MNIST~\citep{clanuwat2018deep}, and Rendered-SST2~\citep{socher2013recursive}, covering diverse domains from natural images to histopathologic data and facial expressions.
        We use the Adam Optimizer with a fixed learning rate of 1e-5 over 4000 training steps (batch size=32).
        Only the vision encoder is fine-tuned while the text encoder remains fixed to preserve the open-vocabulary property of the model.
        All fine-tuned models are publicly available on the HuggingFace Model Hub~\footnote{\url{https://huggingface.co/}} and Modelscope~\footnote{\url{https://www.modelscope.cn/home}} for easy access and reproducibility.
        In Section~\ref{subsec:clip-vit-models}, we present the performance of fine-tuned single-task CLIP-ViT-B/32 and CLIP-ViT-L/14 models on the eight most commonly used eight image classification tasks.
  \item \textbf{ResNet-50 Models}:
        We fine-tune ResNet-50 models on three scene understanding tasks: segmentation, depth estimation, and normal estimation using the NYUv2 dataset, each with a learning rate of 1e-4 for 40 epochs, the learning rate is reduced by a factor of 0.5 every 10 epochs.
        The performance of fine-tuned single-task ResNet-50 models on the NYUv2 dataset is shown in Section~\ref{subsec:resnet-50-models}.
  \item \textbf{GPT-2 Models}:
        We fine-tune GPT-2 models on seven text classification tasks from the GLUE benchmark~\citep{wang2018glue} to evaluate their effectiveness in natural language understanding scenarios.
        The fine-tuning process employs the Adam optimizer with a constant learning rate of 5e-5 over 3 epochs, following standard practices for transformer-based language models.
        These tasks include CoLA, MNLI, MRPC, QNLI, QQP, RTE, and SST-2 .
        The performance evaluation results of these fine-tuned single-task GPT-2 models are presented in Section~\ref{subsec:gpt-2-models}.
  \item \textbf{Flan-T5 Models}:
        We fine-tune both Flan-T5-base and Flan-T5-large models on eight text-to-text generation tasks from the GLUE benchmark.
        For reproducibility and transparency, we provide the detailed prompt templates used for each task in Section~\ref{appendix:prompt_templates}, ensuring that researchers can replicate our experimental setup and build upon our work.
        The performance evaluation results of these LoRA fine-tuned Flan-T5-base and Flan-T5-large models across all eight tasks are presented in Section~\ref{subsec:flan-t5-models}.
\end{enumerate}

Based on the performance metrics detailed in Section~\ref{subsec:clip-vit-models}, \ref{subsec:resnet-50-models}, \ref{subsec:gpt-2-models} and \ref{subsec:flan-t5-models}, we observe that the fine-tuned models demonstrate high accuracy on specific tasks. This observation holds true across various model architectures and task domains, indicating the effectiveness of the fine-tuning process in adapting pre-trained models to excel in particular applications.
Furthermore, fine-tuning a model on one task can lead to both positive or negative transfer effects on other tasks.
Positive transfer occurs when the knowledge gained from the target task enhances the model's performance on another task, while negative transfer arises when the fine-tuning process on the target task hinders the model's ability to perform on others.

\subsection{CLIP-ViT Models}
\label{subsec:clip-vit-models}

\begin{table}[h]
  \caption{Performance of fine-tuned single-task CLIP-ViT-B/32 models on the eight image classification tasks.}
  \label{table:single-task_performance_clip-vit-b-32}
  \setlength{\tabcolsep}{2pt}
  \centering
  \begin{tabular}{lcccccccc}
    \toprule
    \textbf{Model} & \textbf{SUN397} & \textbf{Cars} & \textbf{RESISC45} & \textbf{EuroSAT} & \textbf{SVHN} & \textbf{GTSRB} & \textbf{MNIST} & \textbf{DTD}  \\
    \midrule
    Pre-trained    & 63.2            & 59.8          & 60.7              & 46.0             & 31.6          & 32.5           & 48.2           & 43.9          \\
    \midrule
    SUN397         & \textbf{75.0}   & 47.0          & 54.3              & 46.5             & 28.3          & 26.4           & 44.3           & 41.6          \\
    Cars           & 56.6            & \textbf{78.3} & 50.9              & 38.4             & 30.2          & 30.6           & 49.7           & 41.8          \\
    RESISC45       & 52.0            & 47.2          & \textbf{95.2}     & 56.9             & 23.9          & 24.3           & 39.7           & 35.9          \\
    EuroSAT        & 49.0            & 39.9          & 33.5              & \textbf{99.0}    & 11.8          & 22.9           & 33.8           & 35.5          \\
    SVHN           & 40.5            & 36.3          & 18.9              & 9.8              & \textbf{97.3} & 27.3           & 81.8           & 23.2          \\
    GTSRB          & 36.9            & 33.0          & 20.6              & 21.3             & 41.2          & \textbf{98.9}  & 30.9           & 23.9          \\
    MNIST          & 50.3            & 40.0          & 31.3              & 17.7             & 50.1          & 19.3           & \textbf{99.6}  & 30.7          \\
    DTD            & 54.6            & 51.3          & 36.8              & 25.0             & 28.9          & 21.8           & 47.3           & \textbf{79.7} \\
    \bottomrule
  \end{tabular}
\end{table}
\begin{table}[h]
  \caption{Performance of fine-tuned single-task CLIP-ViT-L/14 models on the eight image classification tasks.}
  \label{table:single-task_performance_clip-vit-l-14}
  \setlength{\tabcolsep}{2pt}
  \centering
  \begin{tabular}{lcccccccc}
    \toprule
    \textbf{Model} & \textbf{SUN397} & \textbf{Cars} & \textbf{RESISC45} & \textbf{EuroSAT} & \textbf{SVHN} & \textbf{GTSRB} & \textbf{MNIST} & \textbf{DTD}  \\
    \midrule
    Pre-trained    & 68.3            & 77.8          & 71.0              & 58.9             & 58.4          & 50.6           & 76.4           & 55.5          \\
    \midrule
    SUN397         & \textbf{82.8}   & 68.4          & 58.1              & 49.9             & 55.0          & 46.3           & 79.5           & 52.8          \\
    Cars           & 67.8            & \textbf{92.9} & 68.7              & 56.4             & 51.7          & 47.7           & 80.5           & 55.6          \\
    RESISC45       & 65.6            & 69.0          & \textbf{97.4}     & 64.3             & 38.3          & 46.6           & 77.7           & 49.9          \\
    EuroSAT        & 65.2            & 69.0          & 40.6              & \textbf{99.2}    & 33.4          & 45.6           & 73.5           & 47.1          \\
    SVHN           & 66.5            & 69.0          & 54.0              & 19.7             & \textbf{97.9} & 48.7           & 92.2           & 50.1          \\
    GTSRB          & 63.4            & 64.8          & 38.7              & 19.6             & 71.0          & \textbf{99.2}  & 75.1           & 45.8          \\
    MNIST          & 56.1            & 49.8          & 53.5              & 26.6             & 48.2          & 33.1           & \textbf{99.8}  & 47.1          \\
    DTD            & 66.8            & 75.3          & 65.5              & 43.7             & 49.5          & 45.0           & 68.5           & \textbf{85.5} \\
    \bottomrule
  \end{tabular}
\end{table}

\begin{figure}[h]
  \centering
  \begin{subfigure}{0.45\textwidth}
    \includegraphics[width=\textwidth]{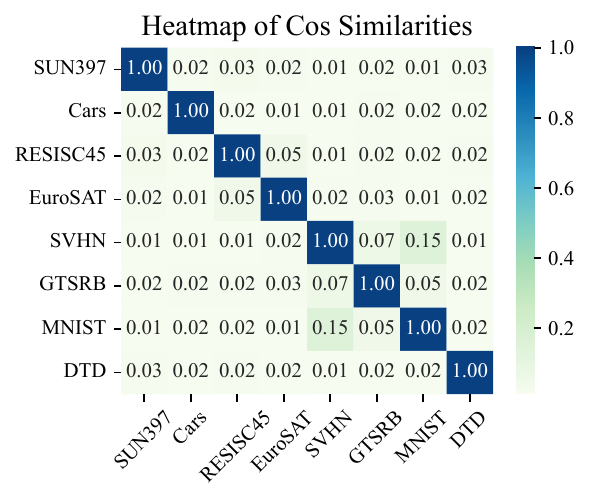}
    \caption{CLIP-ViT-B/32 models.}
    \label{fig:clip-vit-base-patch32_cos}
  \end{subfigure}
  \hspace{0.05\textwidth}
  \begin{subfigure}{0.45\textwidth}
    \includegraphics[width=\textwidth]{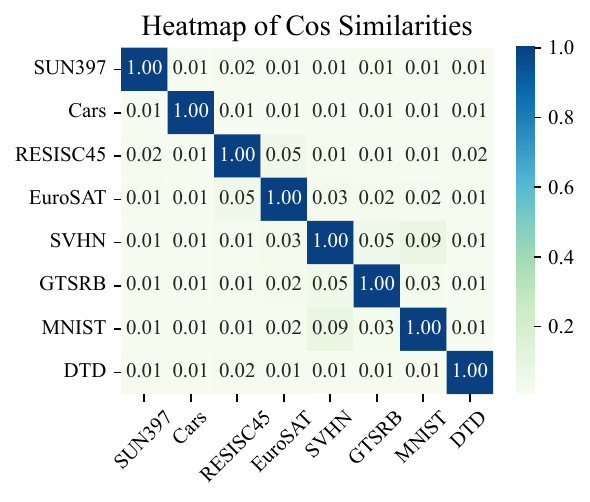}
    \caption{CLIP-ViT-L/14 models.}
    \label{fig:clip-vit-large-patch14_cos}
  \end{subfigure}
  \caption{Cosine similarity of task vectors for CLIP-ViT-B/32 and CLIP-ViT-L/14 models.}
  \label{fig:clip-vit_cos}
\end{figure}

The performance of fine-tuned CLIP-ViT-B/32 and CLIP-ViT-L/14 models on the eight most commonly used image classification tasks is shown in Tables~\ref{table:single-task_performance_clip-vit-b-32} and \ref{table:single-task_performance_clip-vit-l-14}, respectively.
In Figure~\ref{fig:clip-vit_cos}, we visualize the cosine similarity matrices of task vectors for CLIP-ViT-B/32 and CLIP-ViT-L/14 models.
Note that while we present results for the eight core tasks in this paper for conciseness, comprehensive evaluation results for all 20 tasks across CLIP-ViT-B/32, CLIP-ViT-B/16, and CLIP-ViT-L/14 models are available in our online documentation.
Following~\citep{ilharco2022editing}, the task vector is defined as the difference between a fine-tuned model's parameters and the original pre-trained model's parameters: $\tau = \theta_{ft} - \theta_{pre}$, where $\theta_{ft}$ represents the fine-tuned weights and $\theta_{pre}$ represents the pre-trained weights.
We note that the task vectors for models from various tasks are nearly orthogonal. This suggests that the knowledge specific to each task resides in distinct directions or subspaces.
This finding motivates the exploration of locating subspaces in which the knowledge of different tasks can be merged effectively, as discussed in \citet{tang2023concrete,wang2024localizing}.

\subsection{ResNet-50 Models}
\label{subsec:resnet-50-models}

\begin{table}[h]
  \caption{Single-task performance of fine-tuned ResNet-50 models on NYUv2 dataset.}
  \label{table:single-task_performance_nyuv2_resnet50}
  \centering
  \setlength{\tabcolsep}{8pt}
  \small
  \begin{tabular}{lccccc}
    \toprule
    \multirow{2}{*}{\textbf{Method}} & \multicolumn{2}{c|}{\textbf{SEGMENTATION}} & \multicolumn{2}{c|}{\textbf{DEPTH ESTIMATION}} & \textbf{NORMAL}                                                                            \\
                                     & \textbf{mIoU} $\uparrow$                   & \textbf{Pix Acc} $\uparrow$                    & \textbf{Abs Err} $\downarrow$ & \textbf{Rel Err} $\downarrow$ & \textbf{Mean} $\downarrow$ \\
    \midrule
    Segmentation                     & \textbf{52.0}                              & \textbf{73.8}                                  & 242.8                         & 88.7                          & 82.8                       \\
    Depth Estimation                 & 2.3                                        & 6.2                                            & \textbf{42.5}                 & \textbf{17.7}                 & 82.8                       \\
    Normal                           & 2.0                                        & 4.9                                            & 264.0                         & 98.1                          & \textbf{24.7}              \\
    \bottomrule
  \end{tabular}
\end{table}

We fine-tune ResNet-50 models on three distinct scene understanding tasks using the NYUv2 dataset: semantic segmentation, depth estimation, and surface normal estimation.
Each model is trained with a learning rate of 1e-4 for 40 epochs, with the learning rate reduced by a factor of 0.5 every 10 epochs.
Table~\ref{table:single-task_performance_nyuv2_resnet50} presents the single-task performance of these specialized models, where each row represents a model trained exclusively on one task and evaluated across all three tasks using task-specific metrics.
The results demonstrate clear task specialization: the segmentation model achieves optimal performance on segmentation metrics (52.0 mIoU, 73.8 Pixel Accuracy), the depth estimation model excels at depth-related metrics (42.5 Absolute Error, 17.7 Relative Error), and the normal estimation model performs best on surface normal prediction (24.7 Mean angular error).
Notably, models trained on one task show severely poor performance on other tasks, highlighting the challenge of multi-task learning in computer vision applications.

\subsection{GPT-2 Models}
\label{subsec:gpt-2-models}

\begin{table}[h]
  \caption{Performance of fine-tuned GPT-2 models on the seven text classification tasks.}
  \label{table:single-task_performance_gpt-2}
  \centering
  \begin{tabular}{lccccccc}
    \toprule
    \textbf{Model} & \textbf{CoLA} & \textbf{MNLI} & \textbf{MRPC} & \textbf{QNLI} & \textbf{QQP}  & \textbf{RTE}  & \textbf{SST-2} \\
    \midrule
    CoLA           & \textbf{76.8} & 32.8          & 68.4          & 50.4          & 39.2          & 48.0          & 51.0           \\
    MNLI           & 59.5          & \textbf{82.1} & 33.8          & 46.5          & 24.9          & 57.4          & 40.5           \\
    MRPC           & 30.8          & 25.9          & \textbf{80.4} & 47.1          & 65.9          & 49.1          & 49.1           \\
    QNLI           & 58.7          & 38.9          & 30.6          & \textbf{88.3} & 39.9          & 48.7          & 47.0           \\
    QQP            & 31.4          & 25.7          & 62.3          & 45.0          & \textbf{89.6} & 49.1          & 49.1           \\
    RTE            & 52.8          & 47.7          & 37.5          & 53.5          & 33.7          & \textbf{65.3} & 54.9           \\
    SST-2          & 51.8          & 32.9          & 40.2          & 49.8          & 56.8          & 44.4          & \textbf{91.2}  \\
    \bottomrule
  \end{tabular}
\end{table}

We evaluate GPT-2 models fine-tuned on seven distinct text classification tasks from the GLUE benchmark, as detailed in the enumeration above.
Table~\ref{table:single-task_performance_gpt-2} presents the cross-task performance evaluation, where each row represents a model fine-tuned on a specific task (indicated by the row label) and columns show its performance across all seven tasks.
The results demonstrate clear task specialization, with each fine-tuned model achieving its highest performance on its target task (diagonal elements).

\subsection{Flan-T5 Models}
\label{subsec:flan-t5-models}

\begin{table}[h]
  \caption{Performance of LoRA fine-tuned Flan-T5-Base models on the eight text-to-text generation tasks from the GLUE benchmark.}
  \label{table:single-task_performance_flan-t5-base}
  \centering
  \begin{tabular}{lcccccccc}
    \toprule
    \textbf{Model} & \textbf{CoLA} & \textbf{MNLI} & \textbf{MRPC} & \textbf{QNLI} & \textbf{QQP}  & \textbf{RTE}  & \textbf{SST2} & \textbf{STSB} \\
    \midrule
    Pre-trained    & 69.1          & 56.5          & 76.2          & 88.4          & 82.1          & 80.1          & 91.2          & 62.2          \\
    \midrule
    CoLA           & 69.1          & 39.9          & 75.2          & 89.1          & 81.1          & 81.9          & 90.7          & 54.0          \\
    MNLI           & \textbf{69.4} & \textbf{82.7} & 73.8          & 89.3          & 82.0          & 79.4          & 90.9          & 68.1          \\
    MRPC           & 64.0          & 44.9          & \textbf{85.5} & 82.6          & 81.0          & 69.0          & 88.6          & 73.6          \\
    QNLI           & 68.9          & 52.7          & 76.7          & \textbf{90.9} & 82.8          & 79.8          & 91.5          & 68.9          \\
    QQP            & 65.0          & 54.6          & 75.7          & 89.0          & \textbf{84.0} & 81.6          & 90.7          & 75.3          \\
    RTE            & 64.9          & 51.8          & 69.4          & 89.2          & 79.8          & \textbf{84.5} & 90.6          & 70.1          \\
    SST2           & 68.3          & 56.6          & 76.0          & 88.5          & 83.4          & 79.8          & \textbf{92.9} & 62.6          \\
    STSB           & 65.7          & 1.7           & 67.4          & 89.3          & 80.1          & 79.8          & 90.8          & \textbf{87.4} \\
    \bottomrule
  \end{tabular}
\end{table}

\begin{table}[h]
  \caption{Performance of LoRA fine-tuned Flan-T5-Large models on the eight text-to-text generation tasks from the GLUE benchmark.}
  \label{table:single-task_performance_flan-t5-large}
  \centering
  \begin{tabular}{lcccccccc}
    \toprule
    \textbf{Model} & \textbf{CoLA} & \textbf{MNLI} & \textbf{MRPC} & \textbf{QNLI} & \textbf{QQP}  & \textbf{RTE}  & \textbf{SST2} & \textbf{STSB} \\
    \midrule
    Pre-trained    & 73.7          & 56.6          & 82.4          & 91.1          & 85.5          & 85.6          & 94.3          & 87.5          \\
    \midrule
    CoLA           & \textbf{80.2} & 53.9          & 81.4          & 90.8          & 84.5          & 84.1          & 93.9          & 87.1          \\
    MNLI           & 73.7          & \textbf{88.5} & 77.9          & 92.4          & 85.2          & 87.7          & 94.4          & 86.7          \\
    MRPC           & 75.6          & 52.6          & \textbf{89.2} & 92.6          & 84.4          & 86.3          & 94.3          & 86.3          \\
    QNLI           & 73.5          & 54.5          & 82.8          & \textbf{94.4} & 85.8          & 85.2          & 93.7          & 87.1          \\
    QQP            & 74.0          & 53.8          & 82.8          & 92.5          & \textbf{87.2} & 85.6          & 94.5          & 88.3          \\
    RTE            & 75.6          & 57.5          & 69.9          & 92.8          & 83.8          & \textbf{91.7} & 94.6          & 86.0          \\
    SST2           & 73.6          & 55.3          & 82.1          & 91.6          & 85.5          & 85.2          & \textbf{95.2} & 86.9          \\
    STSB           & 73.4          & 39.3          & 82.1          & 92.6          & 86.1          & 83.4          & 94.0          & \textbf{90.9} \\
    \bottomrule
  \end{tabular}
\end{table}

We evaluate Flan-T5-base and Flan-T5-large models fine-tuned using LoRA on eight text-to-text generation tasks from the GLUE benchmark~\citep{wang2018glue}.
Tables~\ref{table:single-task_performance_flan-t5-base} and \ref{table:single-task_performance_flan-t5-large} present the cross-task performance evaluation for both model sizes, where each row represents a model fine-tuned on a specific task and columns show performance across all eight tasks.
The results demonstrate that each specialized model achieves its best performance on its target task (diagonal elements), with the Flan-T5-large models generally outperforming their base counterparts across all tasks.
Notably, when evaluating fine-tuned models on non-target tasks, we observe both positive and negative transfer effects.
For instance, in Flan-T5-base models, the MNLI-specialized model shows positive transfer to QNLI (89.3\% vs 88.\% pre-trained baseline), while the STSB-specialized model exhibits negative transfer to MNLI (1.7\% vs 56.5\% pre-trained baseline).

\section{Experimental Evaluation and Benchmarking Results}
\label{section:evaluation_and_analysis}

In this section, we evaluate the performance of multi-task deep model fusion algorithms on a variety of tasks, as well as analyze the generalization and robustness of these algorithms. We also provide an ablation study to investigate the impact of hyperparameter selection. The majority of experiments presented in this section can be executed using a single NVIDIA RTX 3090 GPU with 24GB memory, ensuring cost-effective reproducibility.

\subsection{Experimental Setup}
\label{section:experimental_setup}

In this section, we conduct a series of multi-task deep model fusion experiments on image classification tasks, text classification tasks, and text-to-text generation tasks to evaluate the performance of multi-task deep model fusion algorithms.
These tasks are chosen to cover a wide range of NLP and CV tasks, as described in Section~\ref{section:multi-task_model_fusion}.

\subsection{Computational Cost Analysis}

While functional requirements listed in Table~\ref{table:implemented_algorithms} differ across methods, computational resource implications also vary significantly.
For instance, ensemble methods incur high inference costs since multiple models must be stored and executed simultaneously.
Model merging methods are generally cost-efficient during inference, as the merged model retains the footprint of a single model, though some approaches require additional data-dependent computations during merging.
Mixing methods typically expand the model size and necessitate partial re-training, leading to substantial computational overhead.

\textbf{Practical Implications:}
For resource-constrained environments, parameter-free methods like parameter average and task arithmetic offer the best cost-performance trade-off.
When additional data is available, alignment-based methods such as RegMean~\citep{jin2022dataless} and RegMean++~\citep{nguyen2025regmean++} provide strong performance with moderate computational overhead.
For applications requiring maximum performance and having sufficient computational resources or access to the test data distribution, test-time adaptive methods like AdaMerging~\citep{yang2023adamerging} or model mixing approaches like WE-MoE~\citep{tang2024merging} are recommended despite their higher computational costs.
Besides, the \texttt{LazyStateDict} utility in FusionBench significantly reduces memory requirements for all methods by enabling on-demand parameter loading, making large-scale model fusion practical.

\subsection{Multi-Task Deep Model Fusion}
\label{section:multi-task_model_fusion}

We start our evaluation by comparing multi-task model fusion algorithms across multiple experimental settings, including both computer vision and natural language processing tasks:

\begin{enumerate}[nolistsep]
  \setlength{\topsep}{0pt}
  \setlength{\itemsep}{0pt}
  \setlength{\parskip}{0pt} \setlength{\parsep}{0pt}
  \item  \textbf{Open-Vocabulary Image Classification}:
        Using the CLIP-ViT models fine-tuned on the eight most commonly used tasks.
        Results for CLIP-ViT-B/32 and CLIP-ViT-L/14 models are presented in Section~\ref{subsubsec:open_vocabulary_image_classification}.
  \item \textbf{Scene Understanding}:
        Using ResNet-50 models on the NYUv2 dataset for segmentation, depth estimation, and normal estimation tasks. Detailed results are provided in Section~\ref{subsubsec:scene_understanding}.
  \item \textbf{Text Classification}:
        Detailed results for GPT-2 models on seven text classification tasks are presented in Section~\ref{subsubsec:text_classification}.
  \item \textbf{Text-to-Text Generation}:
        For Flan-T5-base and Flan-T5-large models fine-tuned with LoRA, the results are presented in Section~\ref{subsubsec:text_to_text_generation}.
\end{enumerate}
In Section~\ref{subsubsec:open_vocabulary_image_classification}, \ref{subsubsec:scene_understanding}, \ref{subsubsec:text_classification} and \ref{subsubsec:text_to_text_generation}, we compare the performance of different multi-task model fusion algorithms across various tasks. Pre-trained models' performance, fine-tuned models' performance, and traditional multi-task learning (MTL) methods are provided for reference. In Appendix~\ref{appendix:single_task_results}, we provide a detailed description of these fine-tuned single-task models.

We have the following key observations:
(1) Multi-task model fusion usually outperforms pre-trained models, showing it can transfer knowledge from multiple single-task models to enhance performance. Pre-trained models lack task-specific knowledge as they are not fine-tuned for downstream tasks.
(2) Adaptive method (AdaMerging) and MoE-based method perform best in multi-task model fusion, showing the effectiveness of adaptive merging and mixture-of-experts approaches.
(3) The performance gap between multi-task model fusion and single-task fine-tuned models (STL) is larger for CLIP-ViT-B/32 compared to CLIP-ViT-L/14. This suggests that multi-task model fusion may be more beneficial for smaller models, as they have more room for improvement through knowledge transfer.
(4) Traditional MTL outperforms most multi-task model fusion methods, which indicates that traditional MTL is still a strong baseline for multi-task learning, and there is room for improvement in multi-task model fusion algorithms.

\subsubsection{Open-Vocabulary Image Classification Using CLIP-ViT Models}
\label{subsubsec:open_vocabulary_image_classification}

\begin{figure}[h]
  \centering
  \includegraphics[width=0.9\linewidth]{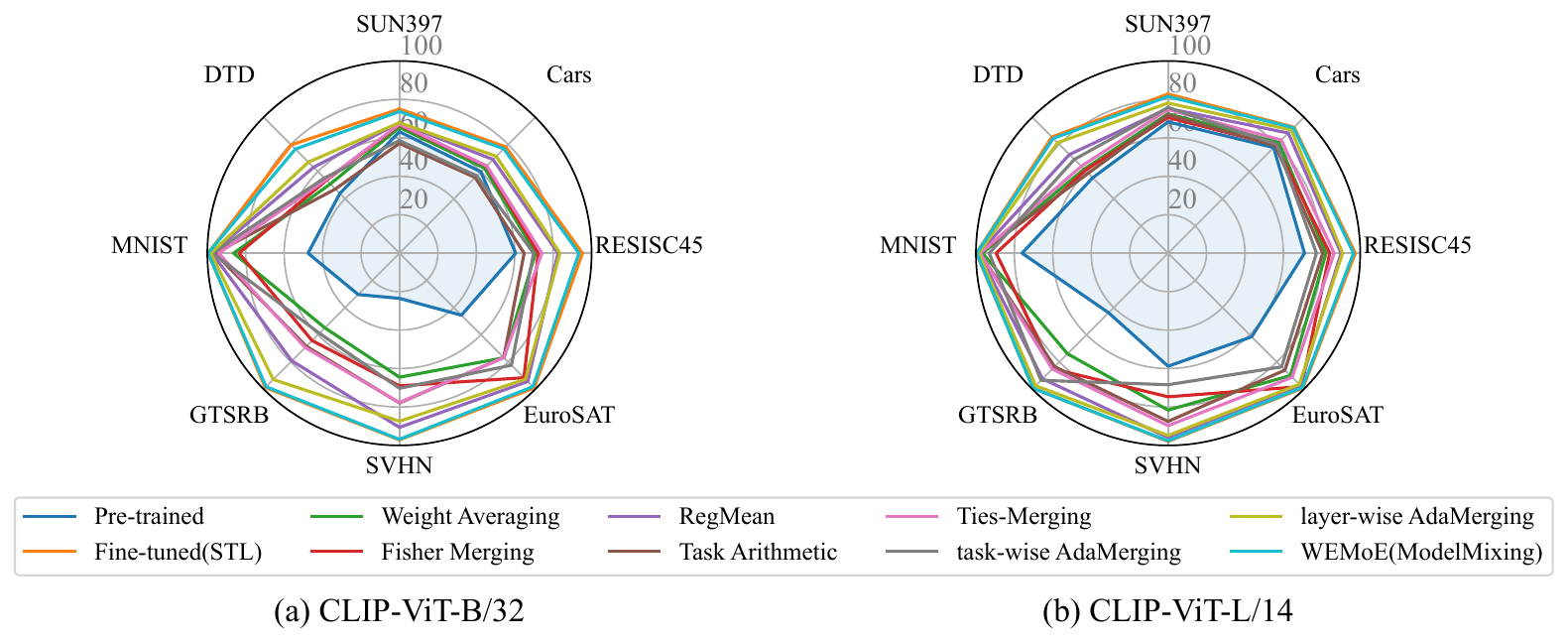}
  \caption{Radar charts comparing the performance of different deep model fusion methods across eight tasks using CLIP-ViT-B/32 and CLIP-ViT-L/14.}
  \label{fig:multi-task_model_fusion}
\end{figure}
\begin{table}[h]
  \caption{Multi-task performance when merging CLIP-ViT-B/32 models on all eight tasks.
    Here \underline{underlined method names} indicate methods that require test-time adaptation training,
    while \textit{italicized method names} indicate model mixing approaches.
  }
  \label{table:multi-task_performance_clip-vit-b-32}
  \begin{center}
    \setlength{\tabcolsep}{1pt}
    \fontsize{9}{12}\selectfont
    \begin{tabular}{lccccccccc}
      \toprule
      \textbf{Method}                           & \textbf{SUN397} & \textbf{Cars} & \textbf{RESISC45} & \textbf{EuroSAT} & \textbf{SVHN} & \textbf{GTSRB} & \textbf{MNIST} & \textbf{DTD} & \textbf{AVG.} \\
      \midrule
      \multicolumn{10}{c}{\textit{Reference Methods}}                                                                                                                                                     \\
      Pre-trained                               & 63.2            & 59.8          & 60.7              & 46.0             & 31.6          & 32.5           & 48.2           & 43.9         & 48.2          \\
      Individual Fine-tuned                     & 75.0            & 78.3          & 95.2              & 99.0             & 97.3          & 98.9           & 99.6           & 79.7         & 90.3          \\
      Traditional MTL                           & 72.3            & 76.6          & 92.2              & 97.9             & 95.5          & 97.7           & 99.3           & 77.7         & 88.6          \\
      \midrule
      \multicolumn{10}{c}{\textit{Multi-Task Model Fusion Methods}}                                                                                                                                       \\
      Weight Averaging                          & 65.4            & 62.6          & 70.8              & 76.9             & 64.5          & 54.9           & 86.3           & 50.9         & 66.5          \\
      Fisher Merging                            & 66.7            & 64.0          & 72.2              & 91.6             & 69.0          & 64.3           & 83.5           & 53.7         & 70.6          \\
      RegMean                                   & 68.6            & 70.0          & 84.6              & 95.4             & 92.6          & 83.4           & 98.4           & 66.1         & 82.4          \\
      RegMean++                                 & 69.3            & 70.5          & 86.7              & 96.1             & 94.1          & 90.4           & 99.0           & 68.7         & 84.4          \\
      Task Arithmetic                           & 57.1            & 55.7          & 64.9              & 76.7             & 77.9          & 68.5           & 96.1           & 47.2         & 68.0          \\
      Ties-Merging                              & 67.1            & 64.2          & 74.1              & 76.8             & 77.7          & 69.4           & 94.1           & 54.0         & 72.2          \\
      \underline{task-wise AdaMerging}          & 58.6            & 56.9          & 69.8              & 82.4             & 70.3          & 58.9           & 97.2           & 55.3         & 68.7          \\
      \underline{layer-wise AdaMerging}         & 67.9            & 71.3          & 83.5              & 92.7             & 87.4          & 92.9           & 98.2           & 67.0         & 82.6          \\
      \underline{\textit{WEMoE (Model Mixing)}} & 73.7            & 76.8          & 93.4              & 98.2             & 96.8          & 98.2           & 99.6           & 76.6         & 89.2          \\
      \textit{SMILE (Model Mixing)}             & 73.6            & 77.8          & 92.0              & 98.3             & 96.9          & 98.1           & 99.6           & 78.1         & 89.3          \\
      \bottomrule
    \end{tabular}
  \end{center}
\end{table}
\begin{table}[h]
  \caption{Multi-task performance when merging CLIP-ViT-L/14 models on all eight tasks.
    Here \underline{underlined method names} indicate methods that require test-time adaptation training,
    while \textit{italicized method names} indicate model mixing approaches.
  }
  \label{table:multi-task_performance_clip-vit-l-14}
  \begin{center}
    \setlength{\tabcolsep}{2pt}
    \fontsize{9}{12}\selectfont
    \begin{tabular}{lccccccccc}
      \toprule
      \textbf{Method}                           & \textbf{SUN397} & \textbf{Cars} & \textbf{RESISC45} & \textbf{EuroSAT} & \textbf{SVHN} & \textbf{GTSRB} & \textbf{MNIST} & \textbf{DTD} & \textbf{Avg.} \\
      \midrule
      \multicolumn{10}{c}{\textit{Reference Methods}}                                                                                                                                                     \\
      Pre-trained                               & 68.3            & 77.8          & 71.0              & 58.9             & 58.4          & 50.6           & 76.4           & 55.5         & 64.6          \\
      Individual Fine-tuned                     & 82.8            & 92.9          & 97.4              & 99.2             & 97.9          & 99.2           & 99.8           & 85.5         & 94.3          \\
      Traditional MTL                           & 79.0            & 89.3          & 94.5              & 98.4             & 96.4          & 98.1           & 99.4           & 83.7         & 92.4          \\
      \midrule
      \multicolumn{10}{c}{\textit{Multi-Task Model Fusion Methods}}                                                                                                                                       \\
      Weight Averaging                          & 72.5            & 81.5          & 82.2              & 90.0             & 81.6          & 74.0           & 96.6           & 61.8         & 80.0          \\
      Fisher Merging                            & 70.6            & 79.4          & 84.1              & 98.1             & 74.7          & 85.0           & 89.5           & 61.0         & 80.3          \\
      RegMean                                   & 76.9            & 89.8          & 93.0              & 97.5             & 96.3          & 94.1           & 98.7           & 77.0         & 90.4          \\
      RegMean++                                 & 77.2            & 89.6          & 92.8              & 97.5             & 96.9          & 96.3           & 99.2           & 78.4         & 91.0          \\
      Task Arithmetic                           & 72.0            & 79.0          & 80.5              & 86.0             & 87.5          & 83.5           & 98.0           & 58.8         & 80.7          \\
      Ties-Merging                              & 74.7            & 83.3          & 86.4              & 91.3             & 89.7          & 85.2           & 97.8           & 63.9         & 84.0          \\
      \underline{task-wise AdaMerging}          & 75.8            & 80.1          & 77.2              & 83.6             & 68.4          & 93.5           & 93.1           & 69.0         & 80.1          \\
      \underline{layer-wise AdaMerging}         & 78.1            & 90.7          & 90.8              & 96.5             & 94.8          & 97.5           & 98.6           & 81.3         & 91.0          \\
      \underline{\textit{WEMoE (Model Mixing)}} & 81.5            & 92.3          & 96.5              & 98.8             & 97.6          & 99.4           & 99.6           & 84.5         & 93.8          \\
      \textit{SMILE (Model Mixing)}             & 81.9            & 92.3          & 95.5              & 99.1             & 98.0          & 98.9           & 99.7           & 83.6         & 93.6          \\
      \bottomrule
    \end{tabular}
  \end{center}
\end{table}

We evaluate model fusion methods on open-vocabulary image classification using CLIP-ViT models fine-tuned on eight diverse image classification tasks spanning natural scenes, satellite imagery, handwritten digits, and texture recognition.
Tables~\ref{table:multi-task_performance_clip-vit-b-32} and ~\ref{table:multi-task_performance_clip-vit-l-14} present comprehensive results for CLIP-ViT-B/32 and CLIP-ViT-L/14 models respectively, while Figure~\ref{fig:multi-task_model_fusion} provides radar chart visualizations highlighting the performance patterns across tasks.
The results demonstrate several key insights:
(1) Model mixing methods (such as WEMoE and SMILE) consistently achieve the best performance, approaching traditional multi-task learning results with averages of 89.2-89.3\% for ViT-B/32 and 93.6-93.8\% for ViT-L/14.
(2) Layer-wise adaptive merging (AdaMerging) significantly outperforms task-wise variants, highlighting the importance of fine-grained parameter-level fusion strategies.
(3) Simple parameter averaging and Fisher merging provide strong baselines but fall short of more sophisticated approaches like RegMean, which leverages data alignment techniques.
(4) The performance gap between fusion methods and individual fine-tuned models is more pronounced for the smaller ViT-B/32 (90.3\% vs 89.3\%) compared to ViT-L/14 (94.3\% vs 93.8\%), suggesting that larger models may be more amenable to effective knowledge combination through fusion techniques.
This phenomenon has also been discussed in the literature on large language model merging~\citep{yadav2024matters}.

\subsubsection{Scene Understanding Using ResNet-50 Models}
\label{subsubsec:scene_understanding}

\begin{figure}[h]
  \centering
  \includegraphics[width=0.8\textwidth]{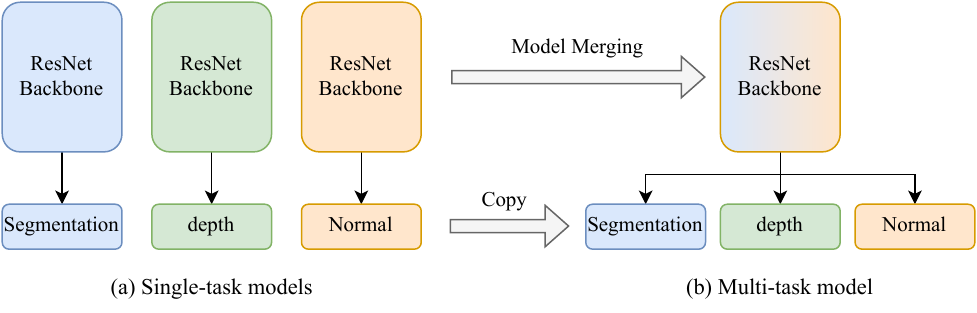}
  \caption{
    Merging ResNet-50 models on three scene understanding tasks: segmentation, depth estimation, and normal estimation.
    Where the backbones are merged and the heads are kept separate.
  }
  \label{fig:nyuv2_resnet50}
\end{figure}
\begin{table}[h]
  \caption{Experimental results of merging single-task Resnet50 models on three NYUv2 tasks.}
  \label{table:nyuv2_resnet50}
  \centering
  \small
  \setlength{\tabcolsep}{5pt}
  \begin{tabular}{c|cc|cc|c}
    \toprule
    \multirow{2}{*}{\textbf{Method}} & \multicolumn{2}{c|}{\textbf{SEGMENTATION}} & \multicolumn{2}{c|}{\textbf{DEPTH ESTIMATION}} & \textbf{NORMAL}                                                                            \\
                                     & \textbf{mIoU} $\uparrow$                   & \textbf{Pix Acc} $\uparrow$                    & \textbf{Abs Err} $\downarrow$ & \textbf{Rel Err} $\downarrow$ & \textbf{Mean} $\downarrow$ \\
    \midrule
    \multicolumn{6}{c}{\textit{Single-Task Learning}}                                                                                                                                                                           \\
    Segmentation                     & \textbf{52.0}                              & \textbf{73.8}                                  & 242.8                         & 88.7                          & 82.8                       \\
    Depth Estimation                 & 2.3                                        & 6.2                                            & \textbf{42.5}                 & \textbf{17.7}                 & 82.8                       \\
    Normal                           & 2.0                                        & 4.9                                            & 264.0                         & 98.1                          & \textbf{24.7}              \\
    \midrule
    \multicolumn{6}{c}{\textit{Multi-Task Model Fusion Methods}}                                                                                                                                                                \\
    Weight Averaging                 & 39.0                                       & 67.0                                           & 55.1                          & 22.7                          & 30.4                       \\
    Task Arithmetic ($\lambda=0.3$)  & 33.6                                       & 63.3                                           & 56.3                          & 23.2                          & 31.3                       \\
    Ties-Merging ($\lambda=0.3$)     & 36.3                                       & 61.7                                           & 60.5                          & 24.5                          & 33.1                       \\
    \bottomrule
  \end{tabular}
\end{table}

Figure~\ref{fig:nyuv2_resnet50} illustrates the multi-task model fusion problem in scene understanding, where three specialized ResNet-50 models trained individually on segmentation, depth estimation, and normal estimation tasks are merged at the backbone level while preserving task-specific heads.
Table~\ref{table:nyuv2_resnet50} presents the results.
The results show that single-task ResNet50 models excel only at their own objectives but fail badly on other tasks, highlighting strong specialization without transferability.
In contrast, model fusion produces balanced performance across tasks: for example, while fusion methods sacrifice some segmentation accuracy compared to the segmentation expert, they dramatically improve depth and normal performance relative to non-specialists.

\begin{table}[h]
  \centering
  \caption{Multi-task performance of GPT-2 models on seven text classification tasks.}
  \label{table:multi-task_performance_gpt-2}
  \begin{tabular}{lcccccccc}
    \toprule
    \textbf{Method}  & \textbf{CoLA} & \textbf{MNLI} & \textbf{MRPC} & \textbf{QNLI} & \textbf{QQP} & \textbf{RTE} & \textbf{SST-2} & \textbf{Avg.} \\
    \midrule
    \multicolumn{9}{c}{\textit{Reference Methods}}                                                                                                  \\
    Fine-tuned (STL) & 76.8          & 82.1          & 80.4          & 88.3          & 89.6         & 65.3         & 91.2           & 82.0          \\
    \midrule
    \multicolumn{9}{c}{\textit{Multi-Task Deep Model Fusion Method}}                                                                                \\
    Simple Average   & 55.0          & 55.1          & 51.0          & 57.6          & 76.7         & 44.8         & 52.5           & 56.1          \\
    Fisher Merging   & 54.8          & 58.0          & 39.5          & 63.3          & 81.5         & 49.1         & 64.7           & 58.7          \\
    RegMean          & 61.7          & 70.4          & 65.4          & 69.7          & 78.8         & 56.0         & 79.7           & 68.8          \\
    Task Arithmetic  & 68.7          & 68.6          & 69.6          & 70.5          & 81.8         & 47.3         & 83.6           & 70.0          \\
    Ties-Merging     & 68.4          & 71.4          & 68.4          & 69.6          & 82.4         & 47.7         & 81.8           & 70.0          \\
    \bottomrule
  \end{tabular}
\end{table}

\subsubsection{Text Classification Using GPT-2 Models}
\label{subsubsec:text_classification}

We evaluate multi-task model fusion methods on natural language understanding tasks using GPT-2 models fine-tuned on seven diverse text classification tasks from the GLUE benchmark~\citep{wang2018glue}.
Table~\ref{table:multi-task_performance_gpt-2} presents the results.

\subsubsection{Text-to-Text Generation Using Flan-T5 Models}
\label{subsubsec:text_to_text_generation}

\begin{table}[h]
  \caption{Experimental results of merging Flan-T5-base (LoRA fine-tuned) models on all eight tasks.
    Here \underline{underlined method names} indicate methods that require test-time adaptation training,
    while \textit{italicized method names} indicate model mixing approaches.
  }
  \label{table:flan-t5-base_lora}
  \small
  \begin{center}
    \setlength{\tabcolsep}{4pt}
    \begin{tabular}{lccccccccc}
      \toprule
      \textbf{Method}                           & \textbf{CoLA} & \textbf{MNLI} & \textbf{MRPC} & \textbf{QNLI} & \textbf{QQP} & \textbf{RTE} & \textbf{SST2} & \textbf{STSB} & \textbf{Avg.} \\
      \midrule
      \multicolumn{10}{c}{\textit{Reference Methods}}                                                                                                                                         \\
      Pre-trained                               & 69.1          & 56.5          & 76.2          & 88.4          & 82.1         & 80.1         & 91.2          & 62.2          & 75.7          \\
      Individual                                & 69.1          & 82.7          & 85.5          & 90.9          & 84.0         & 84.4         & 92.9          & 87.4          & 84.6          \\
      \midrule
      \multicolumn{10}{c}{\textit{Multi-Task Model Fusion Methods}}                                                                                                                           \\
      Weight Averaging                          & 69.7          & 59.7          & 78.9          & 90.1          & 83.8         & 80.5         & 91.2          & 72.0          & 78.2          \\
      Task Arithmetic                           & 68.8          & 55.2          & 78.7          & 89.8          & 83.7         & 79.1         & 91.5          & 72.4          & 77.4          \\
      Ties-Merging                              & 68.3          & 56.3          & 79.4          & 89.8          & 83.7         & 79.4         & 91.6          & 71.2          & 77.5          \\
      \underline{Layer-wise AdaMerging}         & 69.1          & 60.3          & 78.4          & 90.0          & 83.6         & 79.1         & 91.6          & 74.1          & 78.3          \\
      \underline{\textit{SMILE (Model Mixing)}} & 69.3          & 82.9          & 83.8          & 90.6          & 83.9         & 83.4         & 93.1          & 85.1          & 84.0          \\
      \bottomrule
    \end{tabular}
  \end{center}
\end{table}

\begin{table}[h]
  \caption{Experimental results of merging Flan-T5-large (LoRA fine-tuned) models on all eight tasks.}
  \label{table:flan-t5-large_lora}
  \begin{center}
    \setlength{\tabcolsep}{4pt}
    \fontsize{10}{12}\selectfont
    \begin{tabular}{lccccccccc}
      \toprule
      \textbf{Method}                   & \textbf{CoLA} & \textbf{MNLI} & \textbf{MRPC} & \textbf{QNLI} & \textbf{QQP} & \textbf{RTE} & \textbf{SST2} & \textbf{STSB} & \textbf{Avg.} \\
      \midrule
      \multicolumn{10}{c}{\textit{Reference Methods}}                                                                                                                                 \\
      Pre-trained                       & 73.7          & 56.6          & 82.4          & 91.1          & 85.5         & 85.6         & 94.3          & 87.5          & 82.1          \\
      Individual                        & 80.2          & 88.5          & 89.2          & 94.4          & 87.2         & 91.7         & 95.2          & 90.9          & 89.6          \\
      \midrule
      \multicolumn{10}{c}{\textit{Multi-Task Model Fusion Methods}}                                                                                                                   \\
      Weight Averaging                  & 74.6          & 84.3          & 84.1          & 92.8          & 86.3         & 87.4         & 94.8          & 88.0          & 86.5          \\
      Task Arithmetic                   & 76.9          & 85.4          & 85.3          & 93.9          & 85.8         & 88.1         & 95.2          & 87.8          & 87.3          \\
      Ties-Merging                      & 77.1          & 85.1          & 86.3          & 93.9          & 86.0         & 87.7         & 95.1          & 88.0          & 87.4          \\
      \underline{Layer-wise AdaMerging} & 76.7          & 87.6          & 84.8          & 93.8          & 85.9         & 88.1         & 95.2          & 88.6          & 87.6          \\
      \bottomrule
    \end{tabular}
  \end{center}
\end{table}

We extend our evaluation to text-to-text generation tasks using Flan-T5 models fine-tuned with LoRA on eight tasks from the GLUE benchmark~\citep{wang2018glue}, including both classification and regression tasks.
Tables~\ref{table:flan-t5-base_lora} and~\ref{table:flan-t5-large_lora} present results for Flan-T5-base and Flan-T5-large models respectively, demonstrating how model fusion performs in the parameter-efficient fine-tuning paradigm.
The LoRA adapters are merged into the base models before fusion, creating full fine-tuned models that are then subject to various fusion techniques.
Notably, while Flan-T5-base fusion methods show modest improvements over simple averaging (78.2\% vs. individual fine-tuned performance of 84.6\%), the SMILE model mixing method achieves remarkable performance (84.0\%), nearly matching the individual fine-tuned baseline.
For Flan-T5-large, all fusion methods demonstrate stronger performance, with the best methods (Task Arithmetic, Ties-Merging, Layer-wise AdaMerging) achieving 87.3-87.6\% compared to 89.6\% for individual models, suggesting that larger models provide more robust representations for effective knowledge combination across tasks.
This trend aligns with observations from the vision experiments, reinforcing the hypothesis that model scale facilitates more effective parameter-space fusion.

\subsection{Generalization and Robustness Evaluation}
\label{section:generalization_and_robustness_evaluation}

To further assess the generalization and robustness of multi-task model fusion algorithms, we conduct experiments on \textit{unseen tasks} and \textit{corrupted test sets} (or \textit{out-of-distribution test sets}).

From the experimental results in Section~\ref{subsubsec:generalization_experiments} and \ref{subsubsec:robustness_experiments}, we have the following key observations:
(1) The performance of all multi-task model fusion methods on unseen tasks is generally lower than their performance on seen tasks. This is expected, as the models being fused are not explicitly trained on the unseen tasks.
(2) A \textit{negative transfer} is observed in Table~\ref{table:generalization_results_clip-vit-b-32_2} on the RESISC45 dataset, where the merged models exhibit lower accuracy compared to the pre-trained model. The performance of all multi-task model fusion methods on RESISC45 is lower than the pre-trained model, indicating that the knowledge transferred from the seen tasks may not be beneficial or even harmful to this specific unseen task.
(3) The performance of all methods drops significantly on certain types of corruptions, such as pixelation and impulse noise. This highlights the challenge of maintaining robustness under severe distribution shifts and the need for further research in this direction.
(4) When the test distribution is corrupted, adaptive methods may overfit to certain tasks, leading to a decrease in overall performance. This suggests that adaptive methods may need to be further regularized to improve generalization and robustness.

\FloatBarrier
\subsubsection{Generalization Experiments}
\label{subsubsec:generalization_experiments}

\begin{table}[h]
  \caption{Generalization results when merging ViT-B/32 models on six tasks (SUN397, Cars, RESISC45, DTD, SVHN, GTSRB) and evaluating on two unseen tasks (MNIST, EuroSAT).}
  \label{table:generalization_results_clip-vit-b-32}
  \begin{center}
    \setlength{\tabcolsep}{1pt}
    \begin{small}
      \begin{tabular}{l|ccccccc|ccc}
        \toprule
        \multirow{2}{*}{\textbf{Method}} & \multicolumn{7}{c|}{\textbf{Seen Tasks (ACC)}} & \multicolumn{3}{c}{\textbf{Unseen Tasks (ACC)}}                                                                                                                                         \\
                                         & \textbf{SUN397}                                & \textbf{Cars}                                   & \textbf{RESISC45} & \textbf{DTD} & \textbf{SVHN} & \textbf{GTSRB} & \textbf{Avg.} & \textbf{MNIST} & \textbf{EuroSAT} & \textbf{Avg.} \\
        \midrule
        Pre-trained                      & 63.2                                           & 59.9                                            & 60.6              & 43.9         & 23.5          & 30.4           & 46.9          & 47.6           & 45.6             & 46.6          \\
        \midrule
        Fisher Merging                   & 65.5                                           & 67.2                                            & 78.2              & 57.6         & 84.2          & 75.9           & 71.4          & 71.8           & 49.4             & 60.6          \\
        RegMean                          & 68.7                                           & 70.0                                            & 86.5              & 65.9         & 93.9          & 86.7           & 78.6          & 82.2           & 49.3             & 65.7          \\
        Task Arithmetic                  & 64.3                                           & 63.0                                            & 73.2              & 54.9         & 84.7          & 79.5           & 69.9          & 75.5           & 42.6             & 59.1          \\
        Ties-Merging                     & 68.3                                           & 65.5                                            & 76.9              & 54.9         & 75.4          & 72.0           & 68.9          & 73.1           & 47.3             & 60.2          \\
        \underline{AdaMerging}           & 68.4                                           & 71.9                                            & 87.9              & 69.1         & 92.2          & 93.8           & 80.5          & 77.7           & 47.3             & 62.5          \\
        \underline{\textit{WEMoE}}       & 75.4                                           & 77.5                                            & 94.3              & 77.0         & 96.8          & 98.7           & 86.6          & 78.3           & 44.0             & 61.1          \\
        \bottomrule
      \end{tabular}
    \end{small}
  \end{center}
\end{table}
\begin{table}[h]
  \caption{Generalization results when merging ViT-B/32 models on six tasks (SUN397, Cars, GTSRB, EuroSAT, DTD, MNIST) and evaluating on two unseen tasks (RESISC45, SVHN).}
  \label{table:generalization_results_clip-vit-b-32_2}
  \begin{center}
    \setlength{\tabcolsep}{2pt}
    \fontsize{9}{11}\selectfont
    \begin{tabular}{l|ccccccc|ccc}
      \toprule
      \multirow{2}{*}{\textbf{Method}} & \multicolumn{7}{c|}{\textbf{Seen Tasks (ACC)}} & \multicolumn{3}{c}{\textbf{Unseen Tasks (ACC)}}                                                                                                                                         \\
                                       & \textbf{SUN397}                                & \textbf{Cars}                                   & \textbf{GTSRB} & \textbf{EuroSAT} & \textbf{DTD} & \textbf{MNIST} & \textbf{Avg.} & \textbf{RESISC45} & \textbf{SVHN} & \textbf{Avg.} \\
      \midrule
      Pre-trained                      & 63.2                                           & 59.9                                            & 30.4           & 45.6             & 43.9         & 47.6           & 48.4          & 60.6              & 23.5          & 40.1          \\
      \midrule
      Fisher Merging                   & 68.1                                           & 67.4                                            & 67.2           & 86.4             & 58.6         & 81.6           & 71.5          & 60.2              & 42.5          & 51.3          \\
      RegMean                          & 69.4                                           & 70.5                                            & 86.9           & 97.0             & 67.1         & 98.3           & 81.5          & 50.2              & 51.5          & 50.8          \\
      Task Arithmetic                  & 65.2                                           & 63.6                                            & 76.1           & 87.1             & 56.4         & 94.2           & 73.8          & 52.4              & 45.2          & 48.8          \\
      Ties-Merging                     & 68.2                                           & 65.9                                            & 70.0           & 81.2             & 56.0         & 89.0           & 71.7          & 60.3              & 47.3          & 53.8          \\
      \underline{AdaMerging}           & 69.8                                           & 72.4                                            & 95.5           & 95.1             & 70.7         & 98.1           & 83.6          & 48.7              & 60.7          & 54.7          \\
      \underline{\textit{WEMoE}}       & 74.3                                           & 78.1                                            & 98.8           & 98.7             & 75.1         & 99.5           & 87.4          & 47.3              & 51.3          & 49.3          \\
      \bottomrule
    \end{tabular}
  \end{center}
\end{table}

For the generalization experiments, we assess the performance of multi-task deep model fusion algorithms on two unseen tasks after merging ViT-B/32 models trained on six tasks.
The performance of various multi-task model fusion methods, including Fisher Merging~\citep{matena2022merging}, RegMean~\citep{jin2022dataless}, Task Arithmetic~\citep{ilharco2022editing}, Ties-Merging~\citep{yadav2023resolving}, AdaMerging~\citep{yang2023adamerging}, and WEMoE~\citep{tang2024merging}, is compared across both the seen tasks and unseen tasks.
Specifically, we conduct two sets of generalization experiments using the CLIP-ViT-B/32 models:
\begin{itemize}
  \item In the first set, we merge models trained on six tasks (SUN397, Cars, RESISC45, DTD, SVHN, GTSRB) and evaluate the fused model on the unseen tasks (MNIST, EuroSAT). The results are shown in Table~\ref{table:generalization_results_clip-vit-b-32}.
  \item In the second set of experiments, we merge models trained on six tasks (SUN397, Cars, GTSRB, EuroSAT, DTD, MNIST) and evaluate the fused model on the unseen tasks (RESISC45, SVHN). The results are shown in Table~\ref{table:generalization_results_clip-vit-b-32_2}.
\end{itemize}

Tables~\ref{table:generalization_results_clip-vit-b-32} and \ref{table:generalization_results_clip-vit-b-32_2} present the generalization performance of various multi-task model fusion algorithms when merging CLIP-ViT-B/32 models trained on six seen tasks and evaluating their performance on two unseen tasks.
This analysis helps us understand how well the fused models can adapt to new tasks that were not encountered during the training and model fusion process.

\subsubsection{Robustness Experiments}
\label{subsubsec:robustness_experiments}

\begin{table}[h]
  \caption{Ablations of the test data distribution on ViT-B/32 (for all methods, $\lambda=0.3$).
    Here \underline{underlined method names} indicate methods that require test-time adaptation training,
    while \textit{italicized method names} indicate model mixing approaches.
  }
  \label{table:abalation_data_distribution_vit_b_32}
  \begin{center}
    \setlength{\tabcolsep}{2pt}
    \fontsize{8}{9}\selectfont
    \begin{tabular}{l|ccccc|ccccc}
      \toprule
      \textbf{Method}            & \textbf{Cars}                                           & \textbf{EuroSAT}                                          & \textbf{RESISC45} & \textbf{GTSRB} & \textbf{Avg.} & \textbf{Cars} & \textbf{EuroSAT} & \textbf{RESISC45} & \textbf{GTSRB} & \textbf{Avg.} \\
      \midrule
                                 & \multicolumn{5}{c|}{{Clean Test Set}}                   & \multicolumn{5}{c}{{Corrupted Test Set (Motion Blur)}}                                                                                                                                                 \\
      Fisher Merging             & 66.0                                                    & 92.7                                                      & 83.7              & 78.7           & 80.3          & 60.7          & 57.6             & 81.7              & 78.4           & 69.6          \\
      RegMean                    & 72.1                                                    & 97.5                                                      & 88.9              & 93.9           & 88.1          & 70.0          & 71.3             & 87.5              & 86.8           & 78.9          \\
      Task Arithmetic            & 64.6                                                    & 91.8                                                      & 80.2              & 74.8           & 77.9          & 62.4          & 59.2             & 78.5              & 63.3           & 65.9          \\
      Ties-Merging               & 65.2                                                    & 83.3                                                      & 78.1              & 67.4           & 73.5          & 64.4          & 53.9             & 76.4              & 57.1           & 62.9          \\
      \underline{AdaMerging}     & 75.2                                                    & 94.3                                                      & 87.6              & 96.7           & 88.5          & 72.4          & 72.7             & 85.3              & 94.3           & 81.2          \\
      \underline{\textit{WEMoE}} & 77.4                                                    & 98.9                                                      & 94.4              & 99.0           & 92.4          & 76.5          & 74.2             & 93.7              & 97.4           & 85.5          \\
      \midrule
                                 & \multicolumn{5}{c|}{Corrupted Test Set (Impluse Noise)} & \multicolumn{5}{c}{Corrupted Test Set (Gaussian Noise)}                                                                                                                                                \\
      Fisher Merging             & 61.5                                                    & 50.0                                                      & 74.7              & 52.6           & 59.7          & 61.6          & 48.1             & 76.0              & 51.3           & 59.3          \\
      RegMean                    & 66.9                                                    & 51.0                                                      & 80.6              & 68.7           & 66.8          & 69.4          & 41.8             & 84.0              & 67.7           & 65.7          \\
      Task Arithmetic            & 59.8                                                    & 53.3                                                      & 72.3              & 45.0           & 57.6          & 61.5          & 52.5             & 75.0              & 50.1           & 59.8          \\
      Ties-Merging               & 60.2                                                    & 45.6                                                      & 69.8              & 38.3           & 53.5          & 61.8          & 47.3             & 73.1              & 42.3           & 56.1          \\
      \underline{AdaMerging}     & 69.2                                                    & 40.0                                                      & 79.6              & 83.3           & 68.0          & 70.0          & 53.3             & 82.1              & 80.0           & 71.4          \\
      \underline{\textit{WEMoE}} & 75.1                                                    & 9.7                                                       & 91.5              & 91.8           & 67.0          & 76.5          & 9.6              & 92.7              & 88.7           & 66.8          \\
      \midrule
                                 & \multicolumn{5}{c|}{Corrupted Test Set (Pixelate)}      & \multicolumn{5}{c}{Corrupted Test Set (Spatter)}                                                                                                                                                       \\
      Fisher Merging             & 2.2                                                     & 34.0                                                      & 17.0              & 63.2           & 29.1          & 61.4          & 64.2             & 74.6              & 47.3           & 61.9          \\
      RegMean                    & 2.3                                                     & 38.3                                                      & 18.2              & 89.4           & 37.0          & 67.7          & 60.0             & 81.3              & 81.9           & 72.7          \\
      Task Arithmetic            & 2.3                                                     & 33.2                                                      & 19.1              & 65.6           & 30.0          & 61.0          & 62.5             & 72.8              & 57.0           & 63.3          \\
      Ties-Merging               & 3.3                                                     & 31.8                                                      & 18.0              & 58.5           & 27.9          & 61.3          & 52.9             & 70.3              & 48.1           & 58.2          \\
      \underline{AdaMerging}     & 1.3                                                     & 52.9                                                      & 21.0              & 91.0           & 41.5          & 68.4          & 55.9             & 78.3              & 92.3           & 73.7          \\
      \underline{\textit{WEMoE}} & 0.5                                                     & 11.6                                                      & 2.3               & 97.5           & 28.0          & 75.1          & 9.7              & 91.4              & 96.3           & 68.1          \\
      \midrule
                                 & \multicolumn{5}{c|}{Corrupted Test Set (Contrast)}      & \multicolumn{5}{c}{Corrupted Test Set (JPEG Compression)}                                                                                                                                              \\
      Fisher Merging             & 63.8                                                    & 58.4                                                      & 75.5              & 70.4           & 67.0          & 66.3          & 67.6             & 82.6              & 58.9           & 68.8          \\
      RegMean                    & 69.6                                                    & 64.8                                                      & 84.4              & 90.0           & 77.2          & 71.5          & 72.6             & 88.7              & 82.2           & 78.7          \\
      Task Arithmetic            & 62.3                                                    & 55.7                                                      & 75.3              & 70.8           & 66.0          & 63.9          & 66.1             & 80.1              & 61.0           & 67.8          \\
      Ties-Merging               & 64.2                                                    & 52.4                                                      & 74.8              & 63.5           & 63.7          & 65.0          & 59.5             & 77.9              & 53.2           & 63.9          \\
      \underline{AdaMerging}     & 73.1                                                    & 67.4                                                      & 83.0              & 96.2           & 79.9          & 72.9          & 70.7             & 86.3              & 90.6           & 80.1          \\
      \underline{\textit{WEMoE}} & 77.2                                                    & 34.7                                                      & 93.1              & 98.4           & 75.9          & 77.3          & 61.0             & 94.1              & 95.7           & 82.0          \\
      \bottomrule
    \end{tabular}
  \end{center}
\end{table}

To assess the robustness and real-world applicability of multi-task model fusion algorithms, we conduct evaluations on corrupted test sets as presented in Table~\ref{table:abalation_data_distribution_vit_b_32}.
These experiments are crucial for understanding how fusion methods perform under distribution shifts and data quality degradation, which are common challenges in practical deployment scenarios.
We evaluate six different corruption types across four representative image classification tasks (Cars, EuroSAT, RESISC45, GTSRB), including motion blur, impulse noise, Gaussian noise, pixelation, spatter, and JPEG compression.

The results reveal varying degrees of robustness across different fusion methods, with some approaches maintaining reasonable performance under mild corruptions while experiencing significant degradation under severe distortions such as pixelation.
Notably, test-time adaptation-based methods like AdaMerging and WEMoE demonstrate better resilience compared to simpler averaging techniques, though all methods show substantial performance drops under extreme corruption conditions, highlighting the ongoing challenge of maintaining model robustness in multi-task fusion scenarios.

\subsection{Scaling to Large-Scale Neural Networks}
\label{section:applying_model_fusion_to_large_scale_neural_networks}

\begin{table}[h]
  \centering
  \caption{
    Comparison of individual Mistral-7B models and the upscaled model on various benchmark tasks.
    For our method, we set $k_{gate}=8, k=512$, and the total parameter count is 11.2B.
    For a better comparison, we also include the performance of the \texttt{Qwen1.5-14B} model as a reference.
  }
  \label{table:mistral-7b}
  \setlength{\tabcolsep}{2pt}
  \fontsize{10}{12}\selectfont
  \begin{tabular}{ccccc}
    \toprule
    \textbf{Model}                           & \textbf{MMLU}     & \textbf{TruthfulQA} & \textbf{GSM8K}    & \textbf{ARC Challenge} \\
    \midrule
    \texttt{Mistral-7B-v0.1} (pre-trained)   & 59.64             & 42.62               & 38.81             & 53.92                  \\
    \texttt{Qwen1.5-14B} (reference)         & 66.11             & 52.00               & 69.37             & 49.93                  \\
    \midrule
    \texttt{MetaMath-Mistral-7B}             & 60.56             & 44.79               & \textbf{71.49}    & 51.02                  \\
    \texttt{dolphin-2.1-mistral-7b}          & 60.56             & \textbf{55.88}      & 56.93             & 57.00                  \\
    \texttt{speechless-code-mistral-7b-v1.0} & 61.18             & 47.47               & 48.98             & \underline{57.68}      \\
    \midrule
    Simple Average                           & \textbf{61.42}    & 49.95               & 67.40             & 57.59                  \\
    Task Arithmetic ($\lambda=0.3$)          & \underline{61.29} & 49.38               & 66.94             & \textbf{57.94}         \\
    SMILE Upscaled model ($k=512$, 11.2B)    & 60.66             & 52.79               & 67.85             & 54.35                  \\
    SMILE Upscaled model (Dense experts)     & 60.61             & \underline{54.23}   & \underline{70.66} & 55.12                  \\
    \bottomrule
  \end{tabular}
\end{table}

Model fusion methods can also be applied to large-scale neural networks including Large Language Models (LLMs) and Multimodal Large Language Models (MLLMs).
The high computational cost associated with developing LLMs are a significant practical challenge for many researchers.
In FusionBench, we have developed multiple model fusion techniques applicable to LLMs for cheap and efficient model scaling, which can save computational resources for developing new models.
Moreover, FusionBench is actively developing and integrating more LLM and MLLM merging methods, with ongoing work to expand benchmark coverage to large-scale model fusion tasks.
While evaluation methodologies for LLMs and MLLMs are still maturing and computational costs remain a challenge, FusionBench supports established evaluation frameworks such as \texttt{LM-Evaluation-Harness}~\citep{eval-harness} for assessing model performance.
As the toolkit evolves, more comprehensive LLM task evaluations will be included in future releases.

Take merging Mistral-7B models using SMILE upscaling~\citep{tang2024smile} as an example, we compare the performance of the individual Mistral-7B models and the upscaled model on various benchmark tasks in Table~\ref{table:mistral-7b}, as well as the performance of the \texttt{Qwen1.5-14B} model as a reference.
The \texttt{LM-Evaluation-Harness} is utilized to assess the performance of the models.
We fuse three Mistral-7B models, each fine-tuned for a distinct downstream task as showing varying performance across different tasks, thereby incorporating task-specific expertise.
It is observed that the upscaled models and merged methods (Simple Average, Task Arithmetic) generally enhance performance compared to individual models, demonstrating the benefits of model fusion techniques.

\section{Prompt-Based Text-to-Text Generation}
\label{appendix:prompt_templates}

This section details the prompt templates employed for each of the eight text-to-text generation tasks from the GLUE benchmark, see Section~\ref{section:multi-task_model_fusion} for more details.
Within each task, we provide the format of the input text, and the corresponding target text mapping. These templates are crucial in fine-tuning the Flan-T5 models for generating appropriate text outputs tailored to each specific task.

\begin{itemize}
  \item CoLA:
        \begin{itemize}
          \item \textit{Input Text:} "Indicate if the following sentence is grammatically correct or not: "{sentence}". Answer `acceptable' or `unacceptable'."
          \item \textit{Target Text:}
                \begin{itemize}
                  \item 0: "unacceptable"
                  \item 1: "acceptable"
                \end{itemize}
        \end{itemize}

  \item MNLI:
        \begin{itemize}
          \item Input Text: "Does the premise: `{premise}' logically imply, contradict, or is neutral to the hypothesis: `{hypothesis}'? Answer with `entailment', `contradiction', or `neutral'."
          \item Target Text:
                \begin{itemize}
                  \item 0: "entailment"
                  \item 1: "neutral"
                  \item 2: "contradiction"
                \end{itemize}
        \end{itemize}

  \item MRPC:
        \begin{itemize}
          \item \textit{Input Text:} "Are the following sentences `{sentence1}' and `{sentence2}' conveying the same meaning? Answer with `yes' or `no'."
          \item \textit{Target Text:}
                \begin{itemize}
                  \item 0: "no"
                  \item 1: "yes"
                \end{itemize}
        \end{itemize}

  \item QNLI:
        \begin{itemize}
          \item \textit{Input Text:} "Given the context: `{sentence}', does the question `{question}' have an answer based on the information provided? Answer with `yes' or `no'."
          \item \textit{Target Text:}
                \begin{itemize}
                  \item 0: "yes"
                  \item 1: "no"
                \end{itemize}
        \end{itemize}

  \item QQP:
        \begin{itemize}
          \item \textit{Input Text:} "Do the questions `{question1}' and `{question2}' have the same intent? Answer with `yes' or `no'."
          \item \textit{Target Text:}
                \begin{itemize}
                  \item 0: "no"
                  \item 1: "yes"
                \end{itemize}
        \end{itemize}

  \item RTE:
        \begin{itemize}
          \item \textit{Input Text:} "Does the text: `{sentence1}' entail that `{sentence2}' is true? Provide `yes' or `no'."
          \item \textit{Target Text:}
                \begin{itemize}
                  \item 0: "yes"
                  \item 1: "no"
                \end{itemize}
        \end{itemize}

  \item SST-2:
        \begin{itemize}
          \item \textit{Input Text:} "Given the sentence `{sentence}', determine the sentiment. Is it positive or negative?"
          \item \textit{Target Text:}
                \begin{itemize}
                  \item 0: "negative"
                  \item 1: "positive"
                \end{itemize}
        \end{itemize}

  \item STSB:
        \begin{itemize}
          \item \textit{Input Text:} "Consider the sentences `{sentence1}' and `{sentence2}'. On a scale from 1 (completely different) to 5 (completely similar), rate the similarity."
          \item \textit{Target Text:} "{:.1f}", parse to float with one decimal place
        \end{itemize}
\end{itemize}

\paragraph{Reporting Metrics:}
We report accuracy for all tasks except for STSB, where we use Spearman's $\rho$ as the evaluation metric.
For task STSB, the model is expected to output a numerical value. An example from the STSB task is as follows:
\begin{itemize}
  \item \textit{Input}:
        \begin{itemize}
          \item Sentence 1: A plane is taking off.
          \item Sentence 2: An air plane is taking off.
        \end{itemize}
  \item \textit{Output}:
        \begin{itemize}
          \item label: 5
        \end{itemize}
\end{itemize}
We try to parse the output as a numerical value. If the model outputs a numerical value, we can calculate the Spearman's rho between the predicted numerical value and the ground truth numerical value. If the model outputs a non-numerical value, we assume the Spearman's rho is 0, indicating that there is no discernible monotonic increasing or decreasing relationship between the model's predictions and the ground truth. This is a conservative approach, as even non-numerical outputs might contain some relevant information that's being discarded in this evaluation.

\bibliography{25-1243}

\end{document}